\documentclass[a4paper,fleqn]{cas-dc}
\usepackage{natbib}

\usepackage[noend]{algpseudocode}
\usepackage{algorithmicx,algorithm}
\usepackage{graphicx}
\usepackage{pgfplots}
\usepackage{subfigure}
\usepackage{epstopdf}
\usepackage{sidecap}
\usepackage{amsmath}
\usepackage{amssymb}
\usepackage{float}
\usepackage{color}
\usepackage{multirow}   %
\usepackage{caption}    %
\usepackage{booktabs}
\usepackage{ulem}

\usepackage{booktabs}
\usepackage{multirow}
\usepackage{makecell}
\usepackage[table]{xcolor} %
\graphicspath{{figs/}}  %

\providecommand{\teamsTableFont}{\scriptsize}
\usepackage{tabularx}
\usepackage{colortbl}   %
\usepackage{xcolor}
\usepackage{caption}
\usepackage{array}

\def\tsc#1{\csdef{#1}{\textsc{\lowercase{#1}}\xspace}}
\tsc{WGM}
\tsc{QE}

\begin{document}
\renewcommand{\figurename}{Fig.}  %

\let\WriteBookmarks\relax
\def\floatpagepagefraction{1}
\def\textpagefraction{.001}

\title[mode = title]{\textbf{TEAMS}: \textbf{T}ext-prompted spatiot\textbf{E}mporal dual-he\textbf{A}d \textbf{M}amba \textbf{S}nake}

\shortauthors{Ruicheng~Zhang~et~al.}
\shorttitle{\textbf{TEAMS}: \textbf{T}ext-prompted spatiot\textbf{E}mporal dual-he\textbf{A}d \textbf{M}amba \textbf{S}nake}

\author[1]{Ruicheng Zhang}[style=chinese]
\fnmark[$\dagger$]
\author[1]{Jianhui Lei}[style=chinese]
\fnmark[$\dagger$]
\author[1]{Kaiwen Shen}[style=chinese]
\author[1]{Haowei Guo}[style=chinese]
\author[1]{Jun Zhou}[style=chinese]
\author[2]{Bin Chen}[style=chinese]
\author[1]{Mengtang Li}[style=chinese]
\author[1]{Shen Zhao}[style=chinese, orcid=0000-0002-4698-2658] 
\cormark[1]
\author[3]{Shuo Li}[style=chinese]

\cortext[cor1]{Corresponding author.} 
\nonumnote{$\dagger$ These authors contributed equally to this work. }  %
\address[1]{School of intelligent systems engineering, Sun Yat-sen University, Guangzhou 510006, China}
\address[2]{Affiliated Hangzhou First People’s Hospital, Zhejiang University School of Medicine, Zhejiang, China}
\address[3]{School of Engineering, Case Western Reserve University, Cleveland, USA}
\begin{abstract}[S U M M A R Y]
Deep snake is a promising family of instance segmentation methods that accurately predicts object-level contours, thereby overcoming common pixel-level misclassification issues such as mask cavities and jagged edges in semantic segmentation approaches.
However, existing deep snake methods face challenges in handling complex morphological variations, accurately capturing fine-grained organ details, and correcting base detection errors. 
To mitigate these limitations, we propose a cohesive Text-prompted spatiotEmporal dual-heAd Mamba Snake (TEAMS), a novel vision-language Mamba snake framework with three key innovations:
(1) A Spatiotemporal Snake Evolution Strategy (SSES) is introduced to tackle complex morphological variations by capturing bidirectional spatial dependencies along the snake contour and temporal dynamics across evolution steps in a state space model. 
(2) A Contour Morphology-Aware Mamba (CMAM) is proposed to quantify local contour morphologies to modulate the structured attention mask {in the Mamba2 SSD dual form}, which extends Mamba's capability to perceive the relative importance of its input sequence elements for better delineation of fine-grained organ details.
(3) A Text-prompted Collaborative {Dual-Head} Snake (TCDHS) is designed to incorporate cues from textual {prompts} and transfer the evolved contour information to the base detection head, which enhances the deep snake workflow and mitigates wrong detections.
Comprehensive evaluations on five datasets covering different organs and imaging modalities demonstrate that TEAMS outperforms existing semantic and deep snake segmentation methods (e.g., relative mDice/mBF improvements of 6.9\%/9.1\% in a spinal dataset), underscoring its potential as a reliable tool across diverse medical image segmentation scenarios.
\end{abstract}


\maketitle %

%
\section{Introduction}
\label{intro}
\begin{figure}
\centering
\includegraphics[width= 0.5\textwidth]{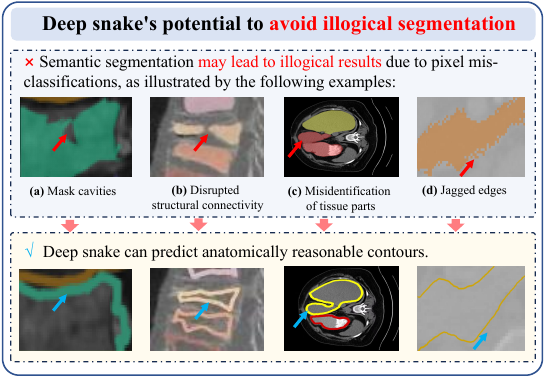}
\caption{Deep snake has the potential to address illogical errors in semantic segmentation caused by pixel misclassifications.}  %
\label{fig:a_new}  %
\vspace{-1em}
\end{figure}

Deep snake exhibits potential advantages over \textit{semantic segmentation} by providing topologically consistent contours and avoiding illogical errors. Current semantic segmentation methods (Fig. \ref{fig:a_new}) mainly rely on pixel-wise classification (\cite{20223d_vessel,cao2021swinunet,2025UM-net}). This may overlook object-level concepts (\cite{DconnNet}) and induce pixel misclassifications, thereby causing illogical segmentation such as mask cavities, disrupted structural connectivity, misidentification of tissue parts, and jagged edges (Fig. \ref{fig:a_new}(a$\sim$d)). In contrast, deep snake is a family of \textit{instance segmentation} methods that directly parameterize contour coordinates to predict high-quality contours (\cite{deep_snake}) via a "detection-then-evolution" workflow. It first detects each target organ (\cite{MSDet}) and initializes a snake contour, then iteratively evolves the contour to approximate the target boundary. This flexible workflow yields anatomically reasonable results and naturally reduces the illogical errors of semantic segmentation methods (\cite{deep_snake,SAMSnake}).

\begin{figure}
\centering
\includegraphics[width= 0.5\textwidth]{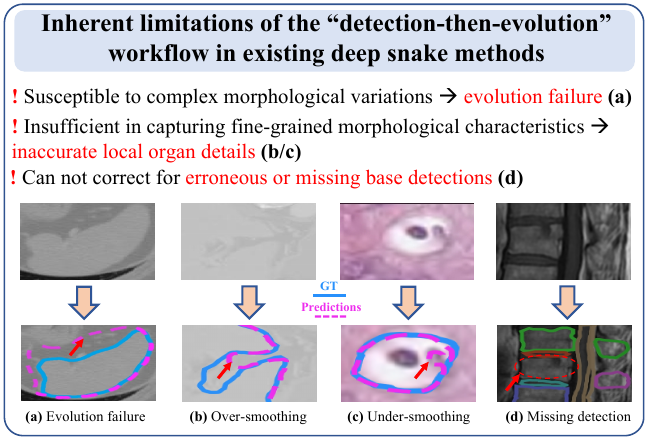}
\caption{Limitations of existing deep snake workflow: susceptibility to morphological variations, inadequate modeling of fine-grained organ details, and inability to correct base detection errors. }
\label{fig:b_new}  %
\vspace{-1em}
\end{figure}

However, existing deep snake methods still struggle {to} comprehensively segment all target organs. The main limitations are: (1) Deep snake evolves points on a structured sequence by predicting their evolution vectors, which is susceptible to complex morphological variations and can result in evolution failure (Fig. \ref{fig:b_new}(a)) (\cite{polyformer}). (2) Current deep snake methods are insufficient in capturing local morphological features, limiting their ability to adjust contour smoothness and accurately delineate organ details (Fig. \ref{fig:b_new}(b)/(c)). (3) The deep snake workflow lacks mechanisms to correct missing or erroneous base detections in its evolution process (Fig. \ref{fig:b_new}(d)). These limitations may lead to suboptimal segmentation when facing challenges such as blurred target boundaries and interference from surrounding structures. Such challenges are common in modern medical image segmentation, particularly in comprehensive segmentation of all tissues within an image. Therefore, deep snake methods require further development.

Mamba has the potential to address the challenges in deep snake methods. As a state space model, Mamba offers a global receptive field with linear complexity, making it suitable for sequence modeling tasks. \textbf{Mamba is inherently well aligned with deep snake methods} because these methods represent target objects as {ordered} contour point sequences. {This better matches Mamba's sequence modeling paradigm and avoids} the risk of fragmenting continuous structural features (\cite{Large_Kernel_mamba}) that commonly arises when integrating Mamba with semantic segmentation in existing work (\cite{vm_net-v2,vmamba,vm_unet,Vision_mamba,vm_unet_h}). However, \textbf{direct integration of Mamba with deep snake methods faces three challenges}:
(1) Mamba’s unidirectional sequence modeling may cause spatiotemporal information loss, as it cannot capture bidirectional dependencies in closed contours or historical evolution {trajectories during snake evolution}. This hinders the accuracy of predicting evolution vectors.
(2) The selective state space modeling in current Mamba blocks is not explicitly conditioned on local contour morphologies (e.g., curvature), limiting its ability to modulate point-wise interactions for accurate delineation of fine-grained organ details.
(3) Although Mamba can facilitate contour evolution, it does not resolve missing or erroneous detections introduced at the base detection stage of the deep snake workflow.

\begin{figure*}
\centering
\includegraphics[width= \textwidth]{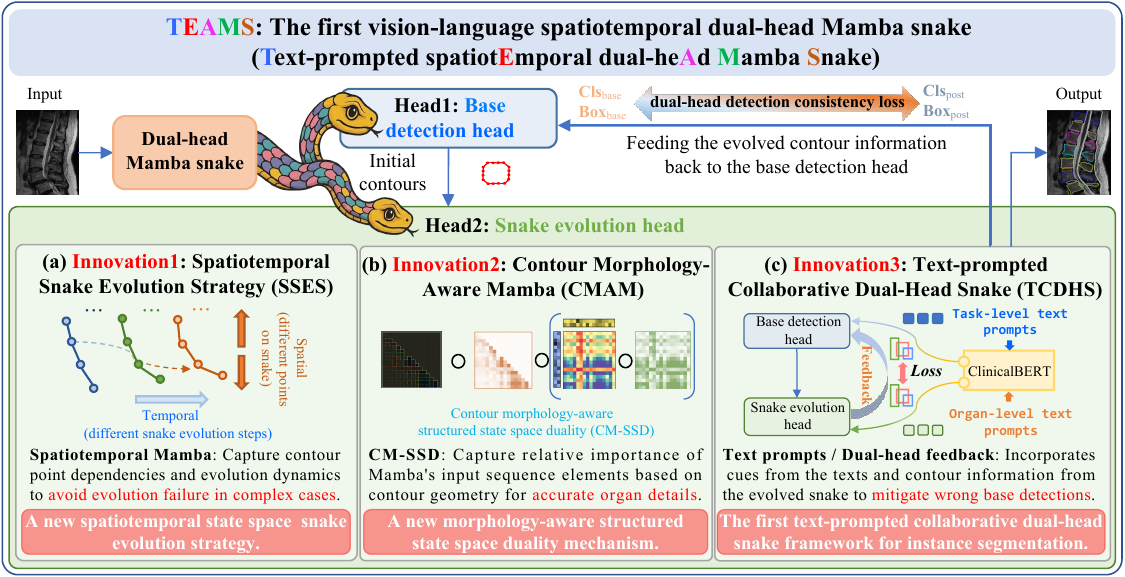}
\caption{Innovation and advantage of our TEAMS: TEAMS is a new deep snake method that captures spatial and temporal context during snake evolution (in SSES), dynamically adapts contour smoothness to fit geometric details (in CMAM), integrates cues from textual {prompts}, and feeds evolved contour features to enhance base detection (in TCDHS) with dual-head consistency feedback.}
\label{fig:c_new}  %
\vspace{-1em}
\end{figure*}

To mitigate these limitations, we propose \textbf{T}ext-prompted spatiot\textbf{E}mporal dual-he\textbf{A}d \textbf{M}amba \textbf{S}nake (TEAMS), the first vision-language Mamba snake framework. {TEAMS introduces a new} \textbf{Spatiotemporal Snake Evolution Strategy} (SSES, Fig. \ref{fig:c_new}(a)) {that formulates snake evolution as bidirectional state space modeling considering historical trajectories, which is further} enhanced by \textbf{{Contour Morphology-Aware Mamba}} (CMAM, Fig. \ref{fig:c_new}(b)) {for more reliable} contour evolution. TEAMS further designs a new \textbf{Text-prompted Collaborative Dual-Head Snake} workflow (TCDHS, Fig. \ref{fig:c_new}(c)) {with vision-language guidance and a dual-head detection consistency feedback mechanism, which leverages textual cues and mitigates wrong detections. These designs equip deep snake methods with new capabilities such as incorporating textual guidance, mitigating wrong detections, and capturing richer spatiotemporal contexts to improve snake evolution, which helps} to robustly segment all organs {under} complex cases (e.g., multiple organs with diverse morphologies, blurred boundaries, and pathological variations).
The {innovations of these} designs are:

(1) SSES (Fig. \ref{fig:c_new}(a) and Fig. \ref{fig2}(b1)): {Different from a simple combination of Mamba and deep snake that uses Mamba to process snake contour point features, SSES processes these features within a newly designed spatiotemporal Mamba architecture that incorporates} bidirectional spatial dependencies among contour points {in the closed snake contour and} trajectory-aware temporal dynamics over historical snake evolution steps. {This enables richer spatiotemporal contexts to} enhance evolution reliability and mitigate evolution failure under complex morphological variations.  %

(2) CMAM (Fig. \ref{fig:c_new}(b) and Fig. \ref{fig2}(b2)): {Unlike a simple combination that directly adopts Mamba, CMAM designs} a contour morphology-aware structured state space duality mechanism {as the basic computational unit in SSES, which exploits} local contour morphologies {to modulate the structured attention mask in} Mamba's dual form. This extends Mamba's capability to perceive the relative importance of its input sequence elements (e.g., the contour points in deep snakes) {and empowers deep snakes to} better {delineate} fine-grained organ details. 

(3) TCDHS (Fig. \ref{fig:c_new}(c) and Fig. \ref{fig2}(b3)): {Different from prior deep snakes that were single-modal and did not take full advantage of the evolved contour information,} TCDHS integrates task-level and organ-level textual prompts {into the “detection-then-evolution” deep snake workflow, allowing it to use textual guidance reflecting} general imaging/anatomical information for detection, and organ shape information for segmentation. {This improves its robustness to local noise and weak boundaries. Furthermore, TCDHS designs a dual-head feedback mechanism that transfers the evolved contour information back to} the base detection head {through the detection consistency loss} between the two heads  (i.e., the base detection head and the snake evolution head). {This enables the deep snake workflow to better exploit the valuable contour-aware cues to mitigate wrong detections.} %

Our contributions are summarized as follows:
\begin{enumerate}
    \item We propose the first vision-language Mamba snake framework to address illogical errors in semantic segmentation.
    \item A spatiotemporal state space snake evolution strategy that jointly models bidirectional spatial dependencies among contour points and historical dynamics across evolution steps is proposed to overcome complex morphological variations.  %
    \item A morphology-aware structured state space duality that exploits local contour morphologies for modulating the structured attention mask is proposed, which extends Mamba's capability to perceive sequence element importance and advances deep snake for better delineation of fine-grained organ details. %
    \item A text-prompted deep snake workflow with dual-head detection consistency is designed to introduce textual cues and enable contour information from the evolved snake to mitigate wrong base detections.

\end{enumerate}

{In this work, we advance our preliminary exploration on {Mamba Snake}} in ACMMM 2025 (\cite{unified}) by introducing a fundamentally new vision-language spatiotemporal dual-head Mamba snake paradigm. {The added contributions are:}

{(1) From the aspect of methodology, TEAMS substantially advances the conference version in terms of both the Mamba snake evolution strategy and the Mamba snake workflow: 
\begin{itemize}
    \item For Mamba snake evolution, TEAMS designs a new spatiotemporal Mamba snake evolution strategy enhanced by a contour morphology-aware structured state space duality. These designs further develop the state space modeling strategy and achieve better snake evolution than the MEB/SSMD in the conference version. %
    \item In the snake workflow, TEAMS introduces a new dual-head feedback mechanism to achieve better detection quality than the DCS. It also introduces textual medical knowledge guidance to alleviate the vulnerability to local noise and weak boundaries, which was absent in the conference version.  %
\end{itemize}

(2) From the aspect of validation,} more challenging data on a wider range of organs and more diverse anatomical variations are used for validation under complex cases. 

{(3) A more comprehensive review on Mamba-based vision models, contour-based segmentation, and vision-language medical models is carried out to provide a panorama of existing work.}

\section{Related Work}
\subsection{Semantic Segmentation Method}
Although semantic segmentation~(\cite{ji2025tgs,huang2026sam,huang2025densely}) has achieved remarkable success, its pixel-wise classification paradigm inherently lacks object-level concepts, which may lead to illogical errors. These methods are primarily developed upon the U-Net architecture~(\cite{compare-unet}) and its variants. For instance, TransUNet~(\cite{2021transunet}) adopts a hybrid CNN-Transformer design to complement local features with global contexts, whereas UNETR~(\cite{unetr}) extends the U-Net framework by adopting a Transformer-based encoder. Semantic segmentation has been extensively applied to medical image segmentation tasks, including the spine~(\cite{2022spine3d}), heart~(\cite{2022bmanet}), brain tumors~(\cite{2021medical_transformer}), blood vessels~(\cite{20223d_vessel}), and polyps~(\cite{2025UM-net}). Recently, foundation models such as Medical Segment Anything Model (MedSAM) have also been explored in semantic segmentation (\cite{medsam}). Nevertheless, these methods fundamentally rely on pixel-wise classification~(\cite{DconnNet}), where \textbf{wrong classifications of certain pixels would lead to illogical segmentation} (the cases in Fig. \ref{fig:a_new}). This limitation \textbf{motivates} the exploration of alternative approaches, such as snake-based segmentation, that explicitly model object boundaries as a complementary paradigm to semantic segmentation~(\cite{SAMSnake}).

\subsection{{Contour}-based Segmentation Method}
{Contour}-based segmentation methods yield topologically consistent contours and reduce illogical segmentation; however, existing approaches still struggle to handle complex medical images. {Since the core ideas of contour-based methods and the family of snake methods are closely related (they both directly predict contour point coordinates), we review both contour-based~(\cite{polyformer,Polygonalformer}) and snake-based~(\cite{zhang2025marl,zhang2025gamed}) methods together in this section following.} Classical snake models~(\cite{snake}) evolve snake contour to the targets using low-level features such as gradients~(\cite{zhao2017}). {Some deep snake models, such as ADMIRE~(\cite{data_MRAVBCE}) and DARNet~(\cite{darnet}), use deep neural networks to extract image features while retaining the traditional snake evolution paradigm. Another line of research establishes a fully learnable deep snake workflow, f}or example: Deep Snake~(\cite{deep_snake}) deforms contours initialized from bounding boxes {towards} the target object boundaries by predicting evolution vectors. {Building upon this workflow, more recent snake/contour-based methods further focus on optimizing snake initialization and evolution procedures, for example:} DANCE~(\cite{DANCE}) enhances contour {evolution} with attention mechanisms to improve contour quality. CD-Net~(\cite{CD-Net}) leverages a graph convolutional network to model spatial dependencies among contour points {for predicting snake evolution vectors}. E2EC~(\cite{E2EC}) introduces a learnable contour initialization module {and a global contour evolution module enhanced with multi-direction alignment and dynamic matching loss. Progressive Deep Snake~(\cite{snake_zixuan}) provides a multi-step supervision strategy from easy to difficult across evolution steps in the snake evolution procedure.} PolySnake (\cite{PolySnake}) optimizes the contour {evolution} process using a recurrent neural network and introduces a boundary map for auxiliary supervision. {BoundaryFormer~(\cite{Polygonalformer}) uses Transformer attention to model vertex feature interactions and predict evolution vectors. SharpContour~(\cite{SharpContour}) proposes a contour evolution approach that estimates the inner/outer state of each vertex to obtain more accurate contours. HemaContour~(\cite{zheng2025hemacontour}) introduces differentiable snake dynamics and a learnable shape-aware objective with overlap, boundary, and curvature terms for intracerebral hemorrhage segmentation.} SAMSnake (\cite{SAMSnake}) integrates EfficientSAM (\cite{EfficientSAM}) to {improve} contour initialization. \textbf{{However, existing} deep snake methods still face three key limitations}: susceptibility to complex morphological variations, {limited perception of} fine-grained morphological features, and inability to correct wrong detections (cases in Fig. \ref{fig:b_new}). This \textbf{motivates} us to develop deep snake {instance segmentation} by {rethinking} the principles of snake evolution, leveraging multi-modal information, and {designing} post-evolution contour {feedback, which differs from existing single-modal deep snakes that follow a unidirectional detection-to-evolution workflow without fully exploiting evolved contour information.}

\subsection{Mamba Model}
\label{sec:mamba_model}
Mamba~(\cite{mamba,MambaTalk}) is a promising state space model for efficient sequence modeling. It captures long-range dependencies using a state space formulation with linear computational complexity (Eq. \ref{eq:mamba_recurrent}). More recently, Mamba2 (\cite{mamba_v2}) introduces structured state space duality (SSD) and simplifies the state transition matrix $\mathbf{A}$ into a scalar, establishing an explicit connection between structured state space models and attention mechanisms. In Mamba2, the time-recursive representation of Mamba (Eq. \ref{eq:mamba_recurrent}) is reformulated into an equivalent matrix transformation form (Eq. \ref{eq:mamba_convolutional}):

\noindent
\begin{subequations}
\begin{minipage}[c]{0.2\textwidth}
\small %
    \begin{equation}
    \left\{
    \begin{aligned}
        \mathbf{h}_t &= \mathbf{A} \mathbf{h}_{t-1} + \mathbf{B} \mathbf{x}_t, \\
        \mathbf{y}_t &= \mathbf{C} \mathbf{h}_t.  %
    \end{aligned}
    \right.
    \label{eq:mamba_recurrent}
    \end{equation}
\end{minipage}%
\hfill %
\begin{minipage}[c]{0.3\textwidth}
\small %
    \begin{equation}
    \left\{
    \begin{aligned}
        \mathbf{Y} &= \mathbf{L} \circ (\mathbf{C} \mathbf{B}^\top) \mathbf{X}, \\
        L_{t1,t2} &=
        \begin{cases}
            \prod_{t=t2+1}^{t1} A_t, & t1 \geq t2, \\
            0, & t1 < t2.
        \end{cases}
    \end{aligned}
    \right. 
    \label{eq:mamba_convolutional}
    \end{equation}
\end{minipage}
\end{subequations}
Eq. \ref{eq:mamba_recurrent} and \ref{eq:mamba_convolutional} show the basic Mamba architecture. In Eq. \ref{eq:mamba_recurrent}, $\mathbf{x}_t$, $\mathbf{h}_t$, and $\mathbf{y}_t$ denote the input, hidden state, and output at time step $t$, respectively; $\mathbf{A}$, $\mathbf{B}$, and $\mathbf{C}$ are respectively learnable state transition, input, and output matrices. In Eq. \ref{eq:mamba_convolutional}, $\mathbf{X}$ denotes the full input sequence, i.e., $\mathbf{X}$ = \{$\mathbf{x}_0$, $\mathbf{x}_1$, ... $\mathbf{x}_T$\}. The matrix $\mathbf{L}$ is a structured mask, where each entry $L_{t1, t2}$ quantifies the influence of element $\mathbf{x}_{t2}$ on $\mathbf{x}_{t1}$. Under the assumption that the state transition matrix $\mathbf{A}$ is scalar, the entry $L_{t1, t2}$ corresponds to the cumulative product of $\mathbf{A}$ from $t_2+1$ to $t_1$. Thus, $\mathbf{L}$ acts as a structured attention mask by modulating pairwise interaction between sequence elements via Hadamard product $\circ$. Therefore, the matrix formulation Eq. \ref{eq:mamba_convolutional} manifests the duality between state space models and attention mechanisms, known as SSD.

Mamba is inherently better suited for deep snake methods than {the patch-based pipelines in} semantic segmentation, but this integration remains largely unexplored. {Recent Mamba-based vision models mostly use Mamba for patch-based pipelines,} where 2D visual features are often {partitioned, flattened, and scanned} into 1D token sequences so that they can be processed by Mamba’s 1D state space operator (\cite{xing2024segmamba, Large_Kernel_mamba, mamba-unet, U-Mamba}){, for example: U-Mamba~(\cite{U-Mamba}) embeds Mamba into a U-Net semantic segmentation network by flattening the spatial dimensions of image features into a one-dimensional sequence and feeding the sequence into Mamba blocks; similarly, Mamba-UNet~(\cite{mamba-unet}) splits the medical image into patches and converts them into a 1D token sequence, then processes them with Mamba blocks arranged in {a U-Net-style encoder-decoder}. Following this pipeline, SegMamba-V2~(\cite{xing2024segmamba}) flattens feature maps into 1D sequences through tri-oriented scanning along coronal, sagittal, and axial planes, and then processes them with Mamba blocks for medical image segmentation. LKM-UNet~(\cite{Large_Kernel_mamba}) flattens image features into pixel-level and patch-level token sequences and then feeds them into bidirectional Mamba modeling. Swin-UMamba~(\cite{Swin-umamba}) further unfolds image patches along four scanning directions into four sequences, and then feeds each sequence into the SSM for enhanced sequence modeling. LoG-VMamba~(\cite{LoG-VMamba}) extracts local and global token sequences and feeds them into Mamba for local-global sequence modeling. EM-Net~(\cite{Em-net}) adopts patch sequence Mamba modeling similar to that in~(\cite{mamba-unet,xing2024segmamba}), and further introduces channel squeeze-reinforce Mamba and frequency-domain learning to improve feature modeling. ShapeMamba-EM~(\cite{Shapemamba-em}) substitutes the self-attention modules in the foundation model image encoder with Mamba layers for efficient feature modeling. In summary, most of these works organize images or feature maps into visual token sequences and process them with the Mamba module~(\cite{U-Mamba,mamba-unet,xing2024segmamba,Large_Kernel_mamba,Swin-umamba,LoG-VMamba,Em-net,Shapemamba-em}); they further introduce spatial scanning strategies~(\cite{xing2024segmamba,Large_Kernel_mamba,Swin-umamba}), enhance patch feature extraction~(\cite{Large_Kernel_mamba,LoG-VMamba,Em-net}), or merge Mamba modules into foundation models~(\cite{Shapemamba-em}). However, such visual token sequence modeling usually relies on flattening or partitioning operations, which may} disrupt continuous spatial features in medical images (\cite{mamba_comprehensive}). {Different from patch-based pipelines,} deep snake methods {\textbf{more naturally match the sequence modeling mechanism of Mamba}} because they represent objects as {ordered contour point sequences by design}. Nevertheless, \textbf{directly applying Mamba to snake evolution may cause spatiotemporal information loss and insufficient modeling of local morphology}, limiting the accuracy of evolution vector prediction and fine-grained boundary delineation. This \textbf{motivates} the design of {TEAMS, a novel Mamba snake workflow with} spatiotemporal state space snake evolution and contour morphology-aware SSD methods tailored to deep snake instance segmentation, {instead of following existing vision Mamba pipelines that partition image features into visual token sequences.}

{
\subsection{{Medical Vision-Language Models}}
\label{sec:Vision_Language_Medical_Models}
Medical vision-language models~(\cite{DeepMolTex}) have recently been used to introduce textual information into medical image analysis. For medical image segmentation, LViT~(\cite{LVit}) designs a vision-language Transformer that constructs text prompts by filling medical annotation information (including infection status, lesion number, and lesion location) associated with each input image into structured text templates; these prompts are then encoded into text features and fused with visual features to guide medical image segmentation. Similarly, RecLMIS~(\cite{RecLMIS}) uses medical notes associated with each individual input image as text prompts; it further introduces conditioned interaction and mutual reconstruction to better align medical notes with visual regions. ATM-Net~(\cite{ATM-Net}) adopts a vision-language U-Net-style framework for lumbar spine segmentation, where anatomy-aware prompts are constructed using annotations that indicate which anatomical structures are present in the image. STPNet~(\cite{STPNet}) constructs a textual prompt library covering lesion distribution, lesion number, and lesion locations; it then retrieves textual features relevant to each image from this library and uses the retrieved textual knowledge to guide segmentation. SemiVL~(\cite{SemiVL}) uses class-specific textual definitions as textual prompts and aligns them with visual features to guide semi-supervised segmentation. With the emergence of medical foundation models, MedCLIP-SAM~(\cite{Medclip-sam}) and MedCLIP-SAMv2~(\cite{Medclip-samv2}) combine medical vision-language representations with SAM to obtain medical segmentation masks through textual prompts. TGSAM-2~(\cite{TGSAM-2}) constructs attribute prior prompts (describing attributes such as relative location, color, and shape) for each organ or lesion to guide medical image segmentation with foundation models. In addition, vision-language models have also been used in medical image analysis tasks such as disease diagnosis~(\cite{Medklip}) and radiotherapy target volume contouring~(\cite{oh2024llm}). These advances \textbf{motivate} us to explore whether textual prompts can provide anatomical and shape cues for deep snake segmentation; our TEAMS contributes a vision-language deep snake workflow in which task-level and organ-level prompts jointly guide the “detection-then-evolution” process. %
}

\section{Methodology}

\begin{figure*}
\centering
\includegraphics[width=1.0 \textwidth]{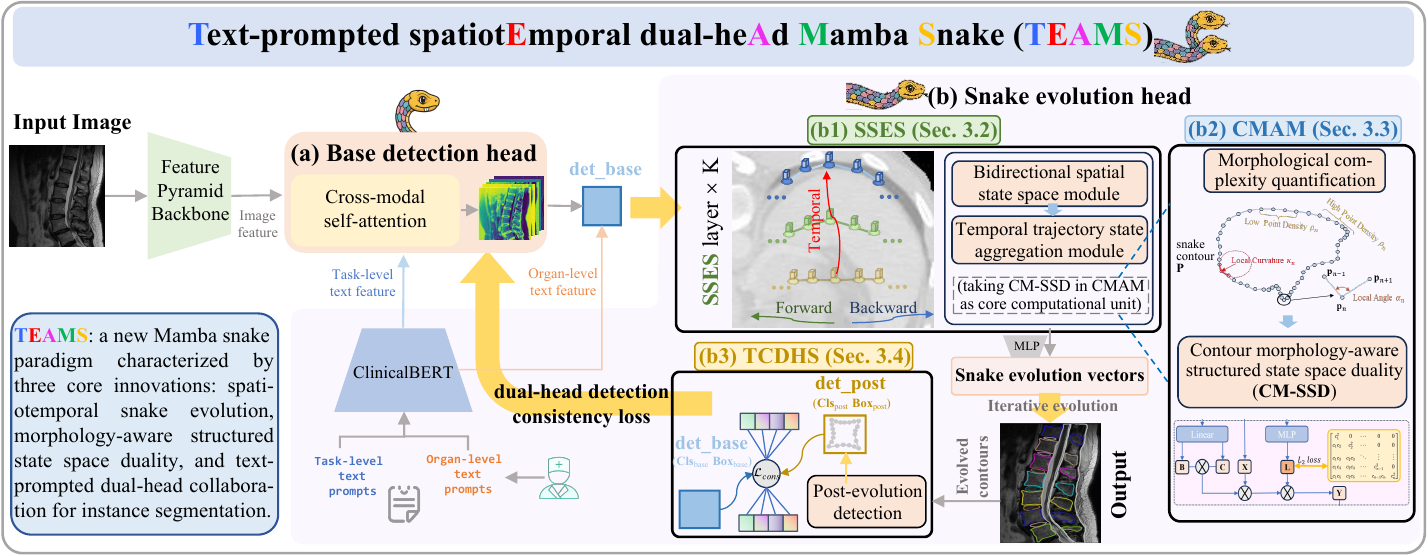}
\caption{The workflow of TEAMS. It first extracts image/text features via a feature pyramid backbone and ClinicalBERT, and fuses them with cross-modal self-attention for base detection (Fig. \ref{fig2}(a)) and contour initialization; it then evolves contours with a dedicated snake evolution head (Fig. \ref{fig2}(b)), where SSES captures spatial and temporal contexts (Fig. \ref{fig2}(b1)) while CMAM adaptively regulates contour smoothness (Fig. \ref{fig2}(b2)) for accurate snake evolution. Finally, TCDHS performs post-evolution detection and designs a dual-head detection consistency loss to align the detections of the base and post-evolution branches (Fig. \ref{fig2}(b3)).
}
\label{fig2}  %
\vspace{-1em}
\end{figure*}

\subsection{Overview}
TEAMS is a cohesive vision-language spatiotemporal dual-head Mamba snake framework. After base detection and contour initialization, TEAMS provides three key designs: 
(1) The \textbf{Spatiotemporal Snake Evolution Strategy} (SSES, Fig. \ref{fig2}(b1), Section \ref{sec:SSES}) formulates snake evolution as state space modeling that includes a \textbf{bidirectional spatial state space module} and a \textbf{temporal trajectory state aggregation module}{, which} jointly models {rich} spatial and temporal contexts to predict snake evolution vectors.  
(2) The \textbf{Contour Morphology-Aware Mamba} (CMAM, Fig. \ref{fig2}(b2), Section \ref{sec:CMAM}) first performs \textbf{morphological complexity quantification} and then constructs the \textbf{contour morphology-aware structured state space duality (CM-SSD)}{, which} introduces morphology-aware modulation in the SSD dual form of Mamba2 {and serves as the basic computational unit in SSES.} 
(3) The \textbf{Text-prompted Collaborative Dual-Head Snake} (TCDHS, Fig. \ref{fig2}(b3), Section \ref{sec:TCDHS}) framework integrates task-level and organ-level text prompts into the snake heads. After contour evolution, TCDHS converts the evolved contours into a contour-aware heatmap and uses it to re-predict {the detections} via a post-evolution detection branch. A \textbf{dual-head detection consistency feedback} {is then designed to align the post-evolution and the base detections, thereby feeding the} contour information back to the base detection head {to improve detection quality}.

\subsection{Spatiotemporal Snake Evolution Strategy}
\label{sec:SSES}
\begin{figure}
\centering
\includegraphics[width= 0.5\textwidth]{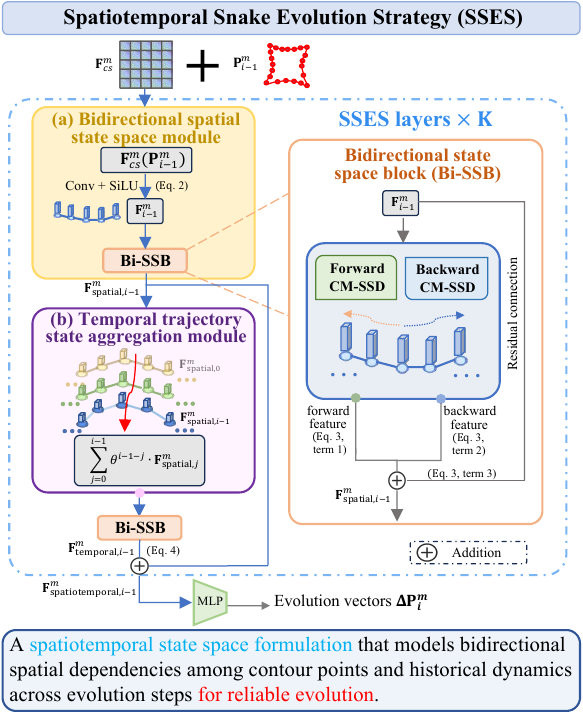}
\caption{The Spatiotemporal Snake Evolution Strategy (SSES). Fig. \ref{fig:SSES}(a) shows the bidirectional spatial state space module. Fig. \ref{fig:SSES}(b) shows the temporal trajectory state aggregation module.
}
\label{fig:SSES}  %
\vspace{-1em}
\end{figure}

The SSES (Fig. \ref{fig:SSES}) iteratively computes the snake contour evolution vectors. For the $m$-th contour in the $i$-th iteration, SSES predicts the evolution vectors $\Delta \mathbf{P}_{i}^{m}$ for the contour point sequence and updates the contour by $\mathbf{P}_{i}^{m} = \mathbf{P}_{i-1}^{m} + \Delta \mathbf{P}_{i}^{m}$. The input to SSES is the organ-level multi-modal feature map $\mathbf{F}_{\text{cs}}^{m}$ and the normalized previous contour $\mathbf{P}_{i-1}^{m}$. SSES formulates the snake contour point sequence by sampling the contour {point} features $\mathbf{F}_{\text{cs}}^{m}(\mathbf{P}_{i-1}^{m})$ and concatenating them with the normalized coordinates $\mathbf{P}_{i-1}^{m}$, forming the sequence input $\mathbf{F}^{m}_{i-1}$. SSES then processes $\mathbf{F}^{m}_{i-1}$ through two complementary branches: \textbf{a bidirectional spatial state space module} (Fig. \ref{fig:SSES}(a)) that processes $\mathbf{F}_{i-1}^{m}$ in forward and backward orders, and \textbf{a temporal trajectory state aggregation module} (Fig. \ref{fig:SSES}(b)) that aggregates historical trajectory {features} over evolution steps. These two branches constitute an SSES layer. Stacking $K$ such SSES layers and applying a multilayer perceptron (MLP) yields $\Delta \mathbf{P}_{i}^{m}$ for the current iteration.
\noindent \textbf{3.2.1 Bidirectional spatial state space module}

This module establishes feature dependencies between adjacent contour points in both forward and backward directions. For the $m$-th organ and $i$-th iteration, the input feature to this spatial branch is calculated by: 
\begin{equation}
\label{eq:input_to_CMAM}
\mathbf{F}_{i-1}^{m} = \mathrm{SiLU}(\mathrm{CircConv}([\mathbf{F}_{\text{cs}}^{m}(\mathbf{P}_{i-1}^m), \mathbf{P}_{i-1}^{m}])),
\end{equation}
where $\mathbf{F}_{\text{cs}}^{m}(\mathbf{P}_{i-1}^m)$ means sampling the features from the organ-level multi-modal feature map $\mathbf{F}_{\text{cs}}^{m}$ (detailed in Section \ref{sec:TCDHS}) at the snake contour points $\mathbf{P}_{i-1}^m$. The resulting feature is then concatenated with the contour coordinates, passed to a circular convolution layer (\cite{snake_zixuan}) and a SiLU layer to form $\mathbf{F}_{i-1}^{m} \in \mathbb{R}^{N \times D}$, which represents the features of the $N$ sequentially {ordered snake contour points}. $D$ means the feature channel number and is set as 128 in our work. $\mathbf{F}_{i-1}^{m}$ is then fed to the bidirectional state space block with a shared residual connection (Bi-SSB): 
\begin{small}
\begin{equation}
\begin{aligned}
\label{eq:Spatial_BI-SSB}
&\mathbf{F}_{\text{spatial},i-1}^{m} 
= \text{Bi-SSB}(\mathbf{F}_{i-1}^{m}) \\
&= \underbrace{\mathrm{CM\text{-}SSD}\left(\mathbf{F}_{i-1}^{m}\right)}_{\text{forward}}
+ \underbrace{\text{rev}\Bigl( \mathrm{CM\text{-}SSD}\bigl(\text{rev}(\mathbf{F}_{i-1}^{m})\bigr) \Bigr)}_{\text{backward}}
+ \underbrace{\mathbf{F}_{i-1}^{m}}_{\text{residual}}
\end{aligned}
\end{equation}
\end{small}

\noindent Eq. \ref{eq:Spatial_BI-SSB} shows how the bidirectional spatial contexts are collected. The three terms in Eq. \ref{eq:Spatial_BI-SSB} mean the forward, backward, and residual features. The first two terms are both computed by CM-SSD (detailed in Section \ref{sec:CMAM}), $\mathrm{rev}(\cdot)$ reverses the sequence order (i.e., in the second term, the sequence is firstly reversed, then processed by CM-SSD, and lastly reversed back to be summed up with other terms). 

\noindent \textbf{3.2.2 Temporal trajectory state aggregation module} 

Simultaneously, the temporal branch captures historical dynamics by aggregating evolution trajectories.
At the $i$-th iteration, the corresponding features in the historical trajectory $\{\mathbf{F}_{\text{spatial},0}^m, \mathbf{F}_{\text{spatial},1}^m, ..., \mathbf{F}_{\text{spatial},i-1}^m\}$ are collected to construct a trajectory-aware temporal state by exponentially decayed aggregation. Specifically, the features from the $j$-th iteration are weighted by $\theta^{i-1-j}$ and aggregated into the current representation, which is then passed into the Bi-SSB block: \begin{equation} 
    \begin{aligned} 
    \label{eq:Temporal_state_aggregation} 
        \mathbf{F}_{\text{temporal},i-1}^{m} = \mathrm{Bi\text{-} SSB}\left(\sum_{j=0}^{i-1} \theta^{i-1-j} \cdot \mathbf{F}_{\text{spatial},j}^m\right),
    \end{aligned} 
\end{equation}
where $\theta \in (0, 1)$ is a decay factor (set as 0.5 in our work) that assigns less importance to earlier evolution features. Then, $\mathbf{F}_{\text{temporal},i-1}^m$ is fused with $\mathbf{F}_{\text{spatial},i-1}^m$ via residual connection to obtain $\mathbf{F}_{\text{spatiotemporal},i-1}^m$, which accomplishes a single SSES layer.

\noindent \textbf{3.2.3 Stacking SSES layers for snake evolution prediction}

$K=6$ SSES layers are stacked {and} the output of the final SSES layer is passed through an MLP{, which} adjust{s} dimensionality and predict{s} the final evolution vectors $\Delta \mathbf{P}_{i}^m$. These vectors are added to the current contour $\mathbf{P}_{i-1}^m$ to produce the updated contour $\mathbf{P}_{i}^m$, thereby completing one iteration of evolution.
{In this way, SSES comprehensively models spatiotemporal contexts during the iterative snake evolution process to adaptively capture the most informative cues for evolution vector prediction}, avoiding the spatiotemporal information loss (i.e., missing opposite-side neighborhood constraints) that could arise from a direct application of a vanilla Mamba to contour sequences. This enhances the robustness and alleviat{es} evolution failure in complex cases.

{
\noindent \textbf{3.2.4 Discussions on SSES module design}  %

(1) \textbf{SSES captures more informative cues than existing temporal modeling strategies}. SSES designs a spatiotemporal Mamba aggregation architecture built upon contour morphology-aware structured state space duality (CM-SSD) as its core computational unit, rather than relying on direct feature aggregation (as in standard temporal feature aggregation) or gated recurrent hidden state propagation (as in recurrent modeling). For more details:
\begin{itemize}
    \item Difference from standard temporal feature aggregation: Standard temporal feature aggregation usually aggregates multi-step evolution features by averaging or weighted averaging. In comparison, SSES further introduces Mamba state space aggregation after exponentially weighted aggregation for selective trajectory state propagation. This makes better use of informative trajectory contexts (e.g., contour point features) in the entire evolution history, and suppresses less relevant historical/spatial information, thereby predicting evolution vectors better than simple weighted averaging.  %
    \item Difference from recurrent modeling: Recurrent modeling usually propagates states through recurrent units. In contrast, SSES propagates them through state space blocks built upon the CM-SSD computational units (detailed in Section \ref{sec:CMAM}), providing a more carefully designed state propagation mechanism that allows local contour morphology to regulate information propagation along ordered contour points. This enables SSES to more effectively select and utilize informative historical contexts, thereby improving evolution vector prediction.
\end{itemize}

(2) \textbf{The reasons for choosing exponential decay mechanism for temporal aggregation}. Exponential decay aggregation better fits snake evolution because it explicitly assigns larger weights to more recent contour trajectories in the iterative evolution process. In this process, historical contours have unequal importance to the current snake evolution. Earlier contours are usually farther from the current contour and contain fewer cues, while more recent contours are closer to the current contour and provide stronger guidance for the next evolution step. Therefore, exponential decay is chosen to provide a simple and explicit temporal weighting scheme: less important early contour information receives smaller weights, whereas more important recent information receives larger weights. Compared with this mechanism, attention-based temporal fusion introduces additional attention parameters for learning temporal weights and lacks an explicit recency constraint, so it may overemphasize less reliable early contours and harm evolution vector prediction. Thus, exponential decay aggregation is chosen to explicitly assign larger weights to more recent states without introducing these parameters, thereby improving the reliability and efficiency of temporal aggregation.
}

\subsection{Contour Morphology-Aware Mamba}
\label{sec:CMAM}

\begin{figure}
\centering
\includegraphics[width= 0.5\textwidth]{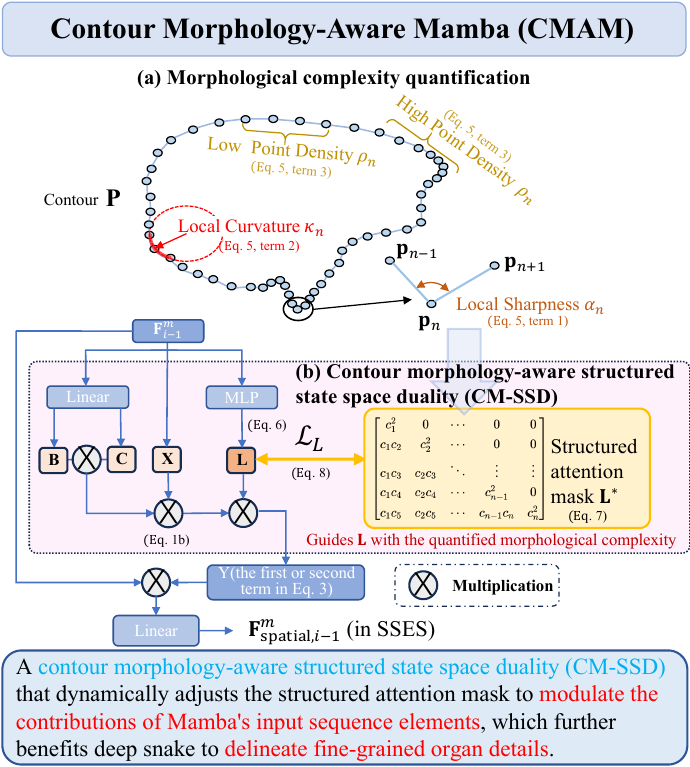}
\caption{The Contour Morphology-Aware Mamba (CMAM). Fig. \ref{CMAM}(a) shows how CMAM quantifies contour morphology complexity. Fig. \ref{CMAM}(b) shows how the CM-SSD is designed to leverage the morphological complexity for modulating the structured attention mask. }
\label{CMAM}  %
\vspace{-2em}
\end{figure}

The CMAM (Fig. \ref{CMAM}) is embedded in the SSES layers to enhance awareness of local geometric details in snake evolution. Given its input features $\mathbf{F}_{i-1}^{m}$, CMAM first calculates the weights $\mathbf{C} \mathbf{B}^\top$ as in the Mamba2 SSD dual form (Eq. \ref{eq:mamba_convolutional}). Furthermore, CMAM performs \textbf{morphological complexity quantification} to indicate the local contour morphologies {as geometric priors}, which is used to dynamically modulate {a learnable} structured attention mask $\mathbf{L}$. This is realized by the \textbf{contour morphology-aware structured state space duality (CM-SSD)}, which {supervises} $\mathbf{L}$ with the {geometric priors} to adaptively modulate the interaction among contour points. CM-SSD serves as the core computational unit within the Bi-SSB in SSES. %

\noindent \textbf{3.3.1 Morphological complexity quantification}

To quantify the local contour morphologies, CMAM introduces three metrics at contour point $\mathbf{p}_n$: local sharpness ($\alpha_n$), local curvature ($\kappa_n$), and point density ($\rho_n$), as shown in Fig. \ref{CMAM}(a). Each metric captures a distinct geometric property and together provide prior information to guide the morphology-aware Mamba modeling:

\begin{equation}
\label{eq:complexity}
\begin{aligned}
h_n 
&= \text{Sigmoid}(w_{\alpha}\alpha_n+w_{\kappa}\kappa_n+w_{\rho}\rho_n) \\
&= \text{Sigmoid} \Biggl( \underbrace{\frac{1}{\pi} \arccos\left(\frac{(\mathbf{p}_n - \mathbf{p}_{n-1}) \cdot (\mathbf{p}_{n+1} - \mathbf{p}_n)}{\| \mathbf{p}_n - \mathbf{p}_{n-1} \| \cdot \| \mathbf{p}_{n+1} - \mathbf{p}_n \|}\right)}_{\text{local sharpness}~(\alpha_n)} \\
&+  \underbrace{\frac{4 \cdot \text{Area}(\mathbf{p}_{n-1}, \mathbf{p}_n, \mathbf{p}_{n+1})}{\| \mathbf{p}_n - \mathbf{p}_{n-1} \| \cdot \| \mathbf{p}_{n+1} - \mathbf{p}_n \| \cdot \| \mathbf{p}_{n+1} - \mathbf{p}_{n-1} \|}}_{\text{local curvature}~(\kappa_n)} \\
&+ \underbrace{\frac{2 d_{\text{avg}}}{\| \mathbf{p}_n - \mathbf{p}_{n-1} \| + \| \mathbf{p}_{n+1} - \mathbf{p}_n \|}}_{\text{point density}~(\rho_n)} \Biggr)
\end{aligned}
\end{equation}

\noindent Eq. \ref{eq:complexity} means that the morphological complexity $h_n$ is calculated by a weighted sum of the three metrics $\alpha_n$, $\kappa_n$, and $\rho_n$ (the weights $w_{\alpha}, w_{\kappa}, w_{\rho}$ are all set to 1). These metrics are respectively calculated by the normalized angle between the vectors $\overrightarrow{\mathbf{p}_{n-1}\mathbf{p}_n}$ and $\overrightarrow{\mathbf{p}_n\mathbf{p}_{n+1}}$, the reciprocal of the circumcircle radius formed by points $(\mathbf{p}_{n-1}, \mathbf{p}_n, \mathbf{p}_{n+1})$, and the inverse of the normalized average point distance between $\mathbf{p}_n$ and its neighbors (where $d_{\text{avg}}$ is the average points distance of all snake contour points).
Since the snake is a closed contour, for the last point $\mathbf{p}_n = \mathbf{p}_{N}$, its next point $\mathbf{p}_{n+1}$ is set as the first point ($\mathbf{p}_1$). Intuitively, higher values of $h_n$ indicate more complex local contour morphologies. %

\noindent \textbf{3.3.2 Contour morphology-aware structured state space duality (CM-SSD)}

The CM-SSD is designed to guide the {learnable} structured attention mask \textbf{L} with the quantified morphological complexity. As illustrated in Fig. \ref{CMAM}(b), we first compute $\mathbf{C} \mathbf{B}^\top$ from the $\mathbf{F}_{i-1}^m$ within the SSD dual form (Eq. \ref{eq:mamba_convolutional}). Then, for each pair of contour points $n_1$ and $n_2$, we construct the structured attention mask $\mathbf{L}$ using the following formulation in an element-wise manner: 
\begin{equation}
\label{eq:predict_L}
    L_{n_1,n_2} =
    \begin{cases}
    \prod_{n=n2+1}^{n1} \text{Sigmoid}\!\left(\phi_f(\mathbf{F}^{m}_{i-1,n})\right) & n_1 \ge n_2\\
    0 & n_1 < n_2
    \end{cases},
\end{equation}  %
where $\phi_f(\mathbf{F}_{i-1,n}^m)$ projects the feature vector of the $n$-th contour point into a scalar to form the structured attention weight $A_n$ in Mamba2 (\cite{mamba_v2}). This formulation is structurally aligned with the structured attention mask $\mathbf{L}$ in Eq.~\ref{eq:mamba_convolutional}, but further clarifies that the $A_n$ is dynamically predicted from input features. The calculated $\mathbf{L}$ is a lower-triangular matrix. In this way, we retain the causal structured attention mask in each directional pass, and obtain bidirectional spatial context via a forward–backward realization (Eq. \ref{eq:Spatial_BI-SSB}). The CM-SSD output is then computed using $\mathbf{Y} = \mathbf{L} \circ (\mathbf{C} \mathbf{B}^\top) \mathbf{F}_{i-1}^m$, where $\mathbf{Y}$ represents the first or second term in Eq. \ref{eq:Spatial_BI-SSB} in SSES.

To further regularize the learning process, we introduce a morphology-aware supervision $\mathbf{L}^*$ for the off-diagonal causal entries in structured attention mask with $n_1 > n_2$, whose elements are computed as:
\begin{equation}
\label{eq:morphology_aware_supervision}
    L_{n_1, n_2}^* = \underbrace{(1 - h_{n_1})}_{\text{susceptibility}} \cdot \underbrace{h_{n_2}}_{\text{influence}} \cdot \underbrace{\exp\left( -\frac{\|\mathbf{p}_{n_1} - \mathbf{p}_{n_2}\|^2}{\sigma^2} \right)}_{\text{distance}},
\end{equation}
Eq. \ref{eq:morphology_aware_supervision} shows how the local contour morphologies are used to guide the structured attention mask in Mamba modeling. In this design, the first term reflects the susceptibility of point $\mathbf{p}_{n_1}$ to external influence, which negatively correlates with the morphological complexity ($h_{n_1}$) at $\mathbf{p}_{n_1}$. The second term represents the influence emitted by $\mathbf{p}_{n_2}$, which is positively correlated with its own morphological complexity ($h_{n_2}$). The third term encodes the Euclidean distance between point $\mathbf{p}_{n_1}$ and $\mathbf{p}_{n_2}$, following an intuition that more distant points exert a weaker influence ($\sigma$ is set as 1). Together, these components enable $L_{n_1, n_2}^*$ to effectively quantify the influence from $n_2$ to $n_1$. After computing $L_{n_1, n_2}^*$, an $L_2$ loss is employed to guide the modulation of the structured attention mask of the SSD mechanism via the following prior supervision:

\begin{equation}
\label{eq:L_L}
\hfill
\mathcal{L}_L = \sum_{n_1 > n_2}
\left\| L_{n_1, n_2}^* - L_{n_1, n_2} \right\|^2,
\hfill\null
\end{equation}

This results in a plug-and-play CM-SSD module that serves as the basic computational unit in the Mamba snake framework and is seamlessly integrated into the SSES layers. Note that Eq. \ref{eq:L_L} is computed for each organ and iteration, however, we omit the organ index $m$ and iteration index $i$ in this section for simplicity.

{
\noindent \textbf{3.3.3 Discussions on CMAM module design}

(1) \textbf{CMAM’s robustness across datasets}. The robustness of CMAM comes from its morphology-guided learnable structured attention mask modeling. In this mechanism, the structured attention mask \textbf{L} used in CMAM is learnable and is dynamically predicted by $\phi_f(\cdot)$ (Eq. \ref{eq:predict_L}), while the three morphology descriptors are used to construct the geometric supervision prior $\mathbf{L}^*$ (Eq. \ref{eq:morphology_aware_supervision}), and to regularize the learning of \textbf{L} through the $L_2$ loss (Eq. \ref{eq:L_L}). Therefore, CMAM does not rely on hand-crafted features for direct decision making; it is a learnable structured mask guided by geometric priors to indicate which contour points have more complex local morphology and should contribute more to other points while being less affected by other points. This provides cross-dataset robustness because:
\begin{itemize}
    \item Learnable mask preserves adaptability across datasets: Since \textbf{L} remains learnable, this guidance does not replace image feature learning. Instead, it preserves the data-driven adaptability of the learnable network, allowing \textbf{L} to robustly adapt to different images, organs, and boundary conditions across datasets.
    \item Morphology guidance improves robustness to complex and blurred boundaries. On top of the learnable mask, $\mathbf{L}^*$ provides contour morphology guidance based on local sharpness, curvature, and point density, which regularizes the structured attention mask toward contour geometry rather than being solely driven by appearance cues that may vary with imaging conditions. This makes the structured attention mask more robust under conditions with insufficient appearance cues (e.g., complex boundary morphologies, blurred boundaries, and noise), enabling it to better delineate fine-grained organ details.
\end{itemize}

(2) \textbf{The suitability of Mamba for snake evolution}. Mamba is naturally more suitable for deep snake evolution than Transformer-based and convolutional alternatives because it better matches the ordered snake contour point sequence and supports flexible feature propagation along the contour. 
\begin{itemize}
    \item Compared with Transformer-based alternatives, Mamba better preserves the ordered structure of snake contour points due to its order-aware state transition mechanism. Since snake contour points are naturally arranged as an ordered sequence, the evolution of each point requires aggregating neighboring point features along the contour, where the contour point order should be explicitly considered. This order-aware feature propagation is naturally supported by the Mamba state transition because the hidden state is sequentially updated along the contour point order and therefore preserves geometric adjacency in the ordered snake contour sequence~(\cite{mamba}). In contrast, this order-aware feature propagation is not naturally achieved by the pairwise attention mechanism in Transformer-based alternatives, which mainly computes similarity-based interactions among input features (i.e., contour point features). Although positional encodings can introduce contour order information~(\cite{shaw2018self}), the similarity-based interaction mechanism is unchanged and does not sequentially propagate features along the contour. Therefore, Mamba better preserves informative neighboring features and suppresses less relevant information along the contour sequence, leading to more accurate evolution vector prediction. 
    \item Compared with CNNs, Mamba supports more flexible contour feature propagation because it adaptively controls the information propagation range. Mamba directly performs selective state space modeling along the contour point sequence, i.e., it can selectively emphasize information from other contour points when computing the feature of each point for evolution vector prediction. For example, in smoother contour regions, Mamba can integrate contour features over a broader neighborhood along the contour, while in sharply curved regions, it can emphasize more local contour point features for evolution vector prediction. In contrast, in CNN-based methods, the same CNN kernels are applied across contour positions, i.e., the same feature aggregation weighted by the CNN kernel is repeated at different positions. This makes CNN-based methods less direct in controlling which neighboring contour information should be propagated or forgotten, limiting their flexibility in determining how far each contour point receives information from its neighborhood according to different morphological complexity.
\end{itemize}

}

\begin{figure*}
\centering
\includegraphics[width=\textwidth]{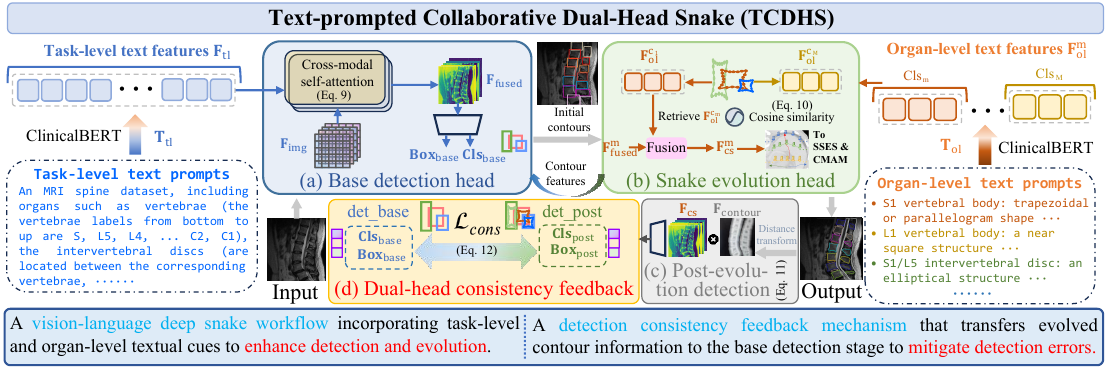}
\caption{The Text-prompted Collaborative Dual-Head Snake (TCDHS). 
In Fig. \ref{fig:tcdhs}(a), task-level text prompts are fused with multi-scale visual features to condition the base detection head, which predicts $\mathbf{Cls}_{\text{base}}$, $\mathbf{Box}_{\text{base}}$, and initializes $\mathbf{P}_0$. In Fig. \ref{fig:tcdhs}(b), organ-level text prompts are aligned with RoI features to form multi-modal feature maps $\mathbf{F}_{\text{cs}}^m$ for snake evolution. In Fig. \ref{fig:tcdhs}(c), the evolved contours $\mathbf{P}$ are converted into contour-aware heatmaps to augment features for a post-evolution detector. In Fig. \ref{fig:tcdhs}(d), a dual-head detection consistency loss is designed to align \{$\mathbf{Cls}_{\text{base}}, \mathbf{Box}_{\text{base}}$\} with \{$\mathbf{Cls}_{\text{post}}, \mathbf{Box}_{\text{post}}$\}, transferring the contour information to the base detector.}
\label{fig:tcdhs}
\vspace{-1em}
\end{figure*}

\subsection{Text-prompted Collaborative Dual-Head Snake}
\label{sec:TCDHS}
TCDHS (Fig. \ref{fig:tcdhs}) contains a \textbf{base detection head}, a \textbf{snake evolution head}, and a \textbf{dual-head detection consistency feedback} to transfer contour shape information from the evolved snake back to the base detector. The base detection head fuses multi-scale visual features with task-level text prompts to predict $\mathbf{Cls}_{\text{base}}$ and $\mathbf{Box}_{\text{base}}$ and initialize the contours $\mathbf{P}_0$. For the $m$-th detected organ, the {snake} evolution head retrieves the corresponding text {prompts} and forms organ-level multi-modal feature map $\mathbf{F}_{\text{cs}}^{m}$, which guides the snake to evolve from $\mathbf{P}_0^m$ to $\mathbf{P}^m$ with the help of SSES and CMAM. Afterwards, all evolved contours $\mathbf{P}$ are converted into contour-aware heatmaps and used to augment features for a post-evolution detector that predicts $\mathbf{Cls}_{\text{post}}$ and $\mathbf{Box}_{\text{post}}$. A dual-head detection consistency loss aligns \{$\mathbf{Cls}_{\text{post}}, \mathbf{Box}_{\text{post}}$\} with \{$\mathbf{Cls}_{\text{base}}, \mathbf{Box}_{\text{base}}$\}, enabling the contour {shape information} to be {perceived by the base detector} for alleviating wrong detections.

\vspace{0.4cm}  %
\noindent \textbf{3.4.1 Base detection head}

The base detection head (Fig. \ref{fig:tcdhs}(a)) first extracts multi-scale visual features from the backbone of a standard detector (e.g., YOLOv8) and fuses them into a unified feature map $\mathbf{F}_{\text{img}}\in\mathbb{R}^{B \times H \times W \times D}$ by resizing and concatenation along the channel dimension. 
It then flattens $\mathbf{F}_{\text{img}}$ along the spatial dimensions into $\mathbf{F}_{\text{img}}^{flat} \in\mathbb{R}^{B \times P \times D}$ ($P=H \times W$). Meanwhile, this head encodes a task-level textual prompt $\mathbf{T}_{\text{tl}}$ using a frozen medical language encoder (e.g., ClinicalBERT), which yields a textual feature vector $\mathbf{f}_{\text{tl}}\in\mathbb{R}^{D}$ providing general task-level imaging and anatomical context across samples (such as imaging modality and the generic topological ordering of the organs, as shown in the leftmost part of Fig. \ref{fig:tcdhs}).
The base detection head then follows prior practice in vision-language feature fusion (\cite{yang2019fast,ye2019Cross}) for image-text fusion, i.e., $\mathbf{f}_{\text{tl}}$ is replicated to $\mathbf{F}_{\text{tl}}\in\mathbb{R}^{B \times P \times D}$ in the first two dimensions, concatenated with $\mathbf{F}_{\text{img}}^{flat}$ in the channel dimension, projected back to channel dimension \textit{D} (forming $\mathbf{F}_{\text{fuse},0}^{flat} \in\mathbb{R}^{B \times P \times D}$), and processed by cross-modal self-attention layers: %

\begin{equation}
\label{eq:task_level_fusion}
\scalebox{0.88}{$\displaystyle
\mathbf{F}_{\text{att},l}^{flat}
=
\operatorname{softmax}\!\left(
\frac{
(\mathbf{F}_{\text{fuse},l-1}^{flat}\mathbf{W}_q)\,
(\mathbf{F}_{\text{fuse},l-1}^{flat}\mathbf{W}_k)^{\top}
}{
\sqrt{d_{att}}
}
\right)
(\mathbf{F}_{\text{fuse},l-1}^{flat}\mathbf{W}_v),
$}
\end{equation}

\noindent where $\mathbf{W}_q,\mathbf{W}_k,\mathbf{W}_v\in\mathbb{R}^{D\times d_{att}}$ are learnable projection matrices, and $d_{att}$ is the attention subspace dimension (set as $d_{att}=D$). 
The $\mathbf{F}_{\text{att},l}^{flat}$ is combined with $\mathbf{F}_{\text{fuse},l-1}^{flat}$ through a residual connection and then passed through a transformer feed-forward operation to obtain $\mathbf{F}_{\text{fuse},l}^{flat}$ following standard transformer architectures~(\cite{vaswani2017attention}), which is fed to the next fusion layer. %
We stack $L=4$ such fusion layers and reshape the output back to the original feature map layout to obtain $\mathbf{F}_{\text{fused}}\in\mathbb{R}^{B \times H \times W \times D}$. $\mathbf{F}_{\text{fused}}$ is fed into the detection head to predict bounding boxes $\mathbf{Box}_{\text{base}}$ and class probabilities $\mathbf{Cls}_{\text{base}}$.
During training, these outputs are supervised with ground truth boxes $\mathbf{Box}^*$ and class labels $\mathbf{Cls}^*$, and optimized using the standard detection loss (denoted as $\mathcal{L}_{\text{base}}$). 
In this way, task-level scene priors are incorporated into the fused features to benefit base detection.

\noindent \textbf{3.4.2 Snake evolution head}

To further enhance snake evolution, the snake evolution head (Fig. \ref{fig:tcdhs}(b)) retrieves the corresponding shape description of each organ from the input organ-level text prompt library. The organ-level {text prompts} cover the general shape of each organ (as shown in the rightmost part of Fig. \ref{fig:tcdhs}){, where each prompt is a sentence describing the anatomical morphology of a specific organ category based on basic medical knowledge}. Following prior practice in (\cite{STPNet}), we retrieve the most relevant organ-level text prompt by matching each Region of Interest (RoI) feature with the most similar entry in the {prompt} library.
Specifically, in the {prompt} library, the organ-level text prompts $\mathbf{T}_{\text{ol}}$ are encoded by ClinicalBERT into a feature matrix and projected to the same channel dimension $D$, denoted as $\mathbf{F}_{\text{ol}} \in \mathbb{R}^{C \times D}$, where each row represents the textual feature of a specific organ category. Meanwhile, for each detection box $\mathbf{Box}_{\text{base}}^m$, its RoI feature is extracted from $\mathbf{F}_{\text{fused}}$ and denoted as $\mathbf{F}_{\text{fused}}^m \in \mathbb{R}^{h \times w \times D}$ (without loss of generality, we assume each RoI is aligned to a fixed size of $h \times w$). Then, the corresponding organ-level text feature $\mathbf{F}^{c_m}_{\text{ol}} \in \mathbb{R}^{D}$ is retrieved from $\mathbf{F}_{\text{ol}}$ based on the cosine similarity between the RoI feature $\mathbf{F}_{\text{fused}}^m$ and each row in $\mathbf{F}_{\text{ol}}$. To enhance this process, we broadcast $\mathbf{F}^{c_m}_{\text{ol}}$ to the same dimensionality as $\mathbf{F}_{\text{fused}}^m$, and then follow (\cite{Luddecke_2022_CVPR}) to minimize the similarity loss: %
\begin{equation}
    \mathcal{L}_{\text{align}} = \frac{1}{M} \sum_{m=1}^{M} \left( 1 - \frac{\phi_v(\mathbf{F}_{\text{fused}}^m) \cdot \phi_t(\mathbf{F}_{\text{ol}}^{c_m})}{\|\phi_v(\mathbf{F}_{\text{fused}}^m)\| \cdot \|\phi_t(\mathbf{F}_{\text{ol}}^{c_m})\|} \right)
    \label{eq:align_loss},
\end{equation}
Eq. \ref{eq:align_loss} promotes the alignment between corresponding image and textual features of the same organ ($\mathbf{F}_{\text{fused}}^m$ and $\mathbf{F}^{c_m}_{\text{ol}}$), where $M$ denotes the number of detected organs, and $\phi_v(\cdot)$ and $\phi_t(\cdot)$ are projection layers that map the flattened RoI~and text features into the same \textit{D}-dimensional embedding space. {This retrieval process is supervised by the ground truth organ labels during training, which align each RoI feature with the textual embedding of its corresponding organ category and thereby facilitate retrieval of the correct organ-level text during inference (\cite{STPNet}).}  %
The aligned features are then fused to form organ-level multi-modal feature maps: $\mathbf{F}_{\text{cs}}^m = [\mathbf{F}_{\text{fused}}^m, \mathbf{F}_{\text{fused}}^m \otimes \mathbf{F}_{\text{ol}}^{c_m}]$, where $\otimes$ denotes the element-wise product to facilitate information fusion across {modalities}  (\cite{chen2020uniter}). Subsequently, $\mathbf{F}_{\text{cs}}^m$ is mapped back to channel dimension $D$ and fed to the SSES in Eq. \ref{eq:input_to_CMAM}{, which} incorporates additional cues about organ shape to $\mathbf{F}_{\text{cs}}^m$ for better {snake} evolution (\cite{rao2022denseclip}).  %

\noindent \textbf{3.4.3 Dual-head detection consistency feedback}

To mitigate missing or erroneous detections, a dual-head detection consistency feedback mechanism is designed to enable the evolved contour information to guide the base detection head. 
Specifically, the evolved contours $\mathbf{P}$ are fed into a post-evolution detector, whose architecture is identical to that of the base detector, but with different input features. 
For the base detection head, the input is $\mathbf{F}_{\text{fused}}$. For the post-evolution detector, the input is augmented with the evolved contour information. Given the evolved contours $\mathbf{P}$, we conduct a distance transform for the contours:
\begin{equation}
    \mathbf{F}_{\text{contour}}(x, y) = \exp \left( - \frac{\min_{\mathbf{p} \in \mathbf{P}} \| (x, y) - \mathbf{p} \|_2^2}{2\sigma^2} \right),
    \label{eq:contour_heatmap}
\end{equation}
\noindent Eq. \ref{eq:contour_heatmap} yields a contour-aware heatmap $\mathbf{F}_{\text{contour}}$ with high values near the contour $\mathbf{P}$, where the numerator means the squared Euclidean distance from an arbitrary point $(x, y)$ on the feature map to the nearest point $\mathbf{p}$ on the evolved contours $\mathbf{P}$, and $\sigma=1$ controls the spread of the Gaussian distribution. %
Subsequently, $\mathbf{F}_{\text{contour}}$ is placed to the original position in the input image {and merged with the detection features}, so that each organ's multi-modal feature map $\mathbf{F}_{\text{cs}}^m$ is weighted by the corresponding $\mathbf{F}_{\text{contour}}^m$ to highlight the features near the contours (Fig. \ref{fig:tcdhs}(c)). The resulting weighted features are then fed into the post-evolution detector, thereby equipping it with contour feature guidance to enhance detection performance. 

Furthermore, we introduce a dual-head detection consistency loss $\mathcal{L}_{\text{cons}}$ (Fig. \ref{fig:tcdhs}(d)) that aligns the predictions of the two detectors:  %
\begin{equation}
    \begin{aligned}
    \mathcal{L}_{cons} &= \mathcal{L}_{\text{cons-cls}} + \mathcal{L}_{\text{cons-box}}\\
    &= \frac{T^2}{M}\sum_{m=1}^{M} \mathcal{L}_\mathrm{KL}\left( \text{sg}\left(\mathbf{Cls}^{(m)}_{\text{post}}\right) ~\Big\|~ \mathbf{Cls}^{(m)}_{\text{base}} \right) \\
    & + \frac{1}{M}\sum_{m=1}^{M} \mathcal{L}_\mathrm{SmoothL1}\left(\mathbf{Box}^{(m)}_{\text{base}},~ \text{sg}\left(\mathbf{Box}^{(m)}_{\text{post}}\right) \right),
    \end{aligned}
    \label{eq:consistency_loss}
\end{equation}
Eq. \ref{eq:consistency_loss} shows that for the $m^{\text{th}}$ detection, its class probability vectors from the post-evolution and base detectors ($\mathbf{Cls}^{(m)}_{\text{post}}$ and $\mathbf{Cls}^{(m)}_{\text{base}}$) are aligned by the KL divergence loss, and its bounding boxes ($\mathbf{Box}^{(m)}_{\text{post}}$ and $\mathbf{Box}^{(m)}_{\text{base}}$) with the $\mathrm{SmoothL1}$ loss, where $M$ represents the number of detected organs. The operator $\mathrm{sg}(\cdot)$ stops gradients from flowing into the post-evolution branch, so that it serves as a teacher to guide the base detector in a one-way manner. $T$ is a weighting factor and is set as 1.
Both detectors are supervised with their respective class labels and bounding box annotations; meanwhile, through the one-way consistency loss (Eq. \ref{eq:consistency_loss}), the contour information from the post-evolution detector is transferred to the base detector for calibrating its class probability prediction and bounding box regression.

\noindent \textbf{3.4.4 Training objective}

With the introduction of TCDHS, the overall training objective of the TEAMS framework is formulated as a weighted sum of multiple loss components:
\begin{equation}
\begin{aligned}
\mathcal{L}_{\mathrm{TEAMS}} &=
\lambda_{\mathrm{base}} \mathcal{L}_{\mathrm{base}} +
\lambda_{\mathrm{evo}} \mathcal{L}_{\mathrm{evo}} +
\lambda_{\mathrm{post}} \mathcal{L}_{\mathrm{post}} \\
&+ \lambda_{L} \mathcal{L}_{L}  +
\lambda_{\mathrm{align}} \mathcal{L}_{\mathrm{align}} +
\lambda_{\mathrm{cons}} \mathcal{L}_{\mathrm{cons}},
\end{aligned}
\end{equation}
where each term corresponds to a specific module in the TEAMS (base detection, contour evolution, post-evolution detection, attention guidance, cross-modal alignment, and consistency supervision). $\mathcal{L}_{\mathrm{base}}$ and $\mathcal{L}_{\mathrm{post}}$ are both adopted from standard YOLO network, and $\mathcal{L}_{\mathrm{evo}}$ is the snake evolution loss adopted from (\cite{deep_snake}). The hyperparameters $\lambda_{(\cdot)}$ balance their relative contributions to ensure the loss components remain at the same order of magnitude, set as $\lambda_{\mathrm{base}}=1$, $\lambda_{\mathrm{evo}}=1.5$, $\lambda_{\mathrm{post}}=1$, $\lambda_{L}=0.1$, $\lambda_{\mathrm{align}}=0.1$, and $\lambda_{\mathrm{cons}}=0.5$.
We adopt a training schedule by first training the base detector and evolution network for $\sim$250 epochs, and then enabling the remaining losses for joint optimization for another $\sim$100 epochs.

{
\noindent \textbf{3.4.5 Discussions on TCDHS module design}

(1) \textbf{The vision-language guidance design}. TCDHS provides vision-language guidance by integrating task-level prompts with global visual features and dynamically retrieved organ-level prompts with region-specific visual features to guide the deep snake workflow. As detailed in Section \ref{sec:Vision_Language_Medical_Models}, in recent vision-language guidance studies, textual prompts are commonly constructed from medical prior knowledge (e.g., anatomical structures, lesion attributes, and category definitions), which are then incorporated into visual feature modeling and influence downstream predictions~(\cite{LVit,RecLMIS,ATM-Net,STPNet,SemiVL,TGSAM-2}). These text prompts can either serve as fixed anatomical (or lesion) priors or be constructed by dynamically filling templates according to the information in each image. For example, SemiVL~(\cite{SemiVL}) uses category-level textual definitions that remain fixed across images of the same category. This inspires the task-level prompts in TEAMS, which are kept identical between training and inference to provide task priors in TEAMS. Meanwhile, STPNet~(\cite{STPNet}) first constructs a fixed prompt library and then retrieves the most relevant prompts for each image. This inspires the design of organ-level text prompts in TEAMS, where organ prompts are dynamically retrieved (rather than using a single fixed prompt for all organs) from a prompt library according to detected regions without requiring manual text input during inference. The alignment between selected organ-level textual cues and visual features is enhanced by the loss in Eq. \ref{eq:align_loss}. In this way, the textual prompts in TEAMS provide vision-language guidance to enhance the performance of the "detection-then-evolution" deep snake workflow.

(2) \textbf{The mechanism of dual-head feedback in mitigating missing detections}. TCDHS mitigates missing detections because the dual-head feedback mechanism is designed to achieve higher detection quality, which naturally mitigates the "completely missing" cases where the snake evolution cannot be initiated. This is determined by the filtering mechanism in the detection pipeline. In this pipeline, if the detection quality of a “hard” true target is low (e.g., the confidence score of the correct class is less than some low threshold such as 0.1), it will be difficult to distinguish from background or false positive proposals and will be filtered out by score thresholding or NMS. This results in cases where the target is "completely missed by the detector". This issue is less critical for snake evolution during training, where ground truth boxes and contours provide reliable initialization and supervision. However, during testing, the “completely missed” targets will harm snake initialization and segmentation. With this in mind, a direct way to mitigate the issue of “completely missed” targets is to improve the detection quality (e.g., improve the confidence score of the correct class so that the targets can be distinguished from background or false positive proposals and are less likely to be filtered out). 

This motivates the dual-head feedback mechanism, which transfers shape information from evolved contours back to the base detection head through the dual-head detection consistency loss. This loss aligns the base detector with the post-evolution detector, so contour-aware predictions from the post-evolution detector can serve as an additional teacher signal. In this way, the base detector is trained to approximate the contour-aware predictions of the post-evolution detector. This encourages the base detector to better capture shape cues from the input features and associate them with correct organ categories and accurate bounding boxes. As a result, the base detector achieves better detection quality for true targets, such as higher confidence scores for the correct classes and more accurate bounding boxes. This enhanced detection quality reduces their chances of being filtered out by score thresholding or NMS due to low confidence or localization errors, thereby mitigating the risk of “completely missed” detections.

}

\begin{figure*}
\centering
\includegraphics[width=\textwidth]{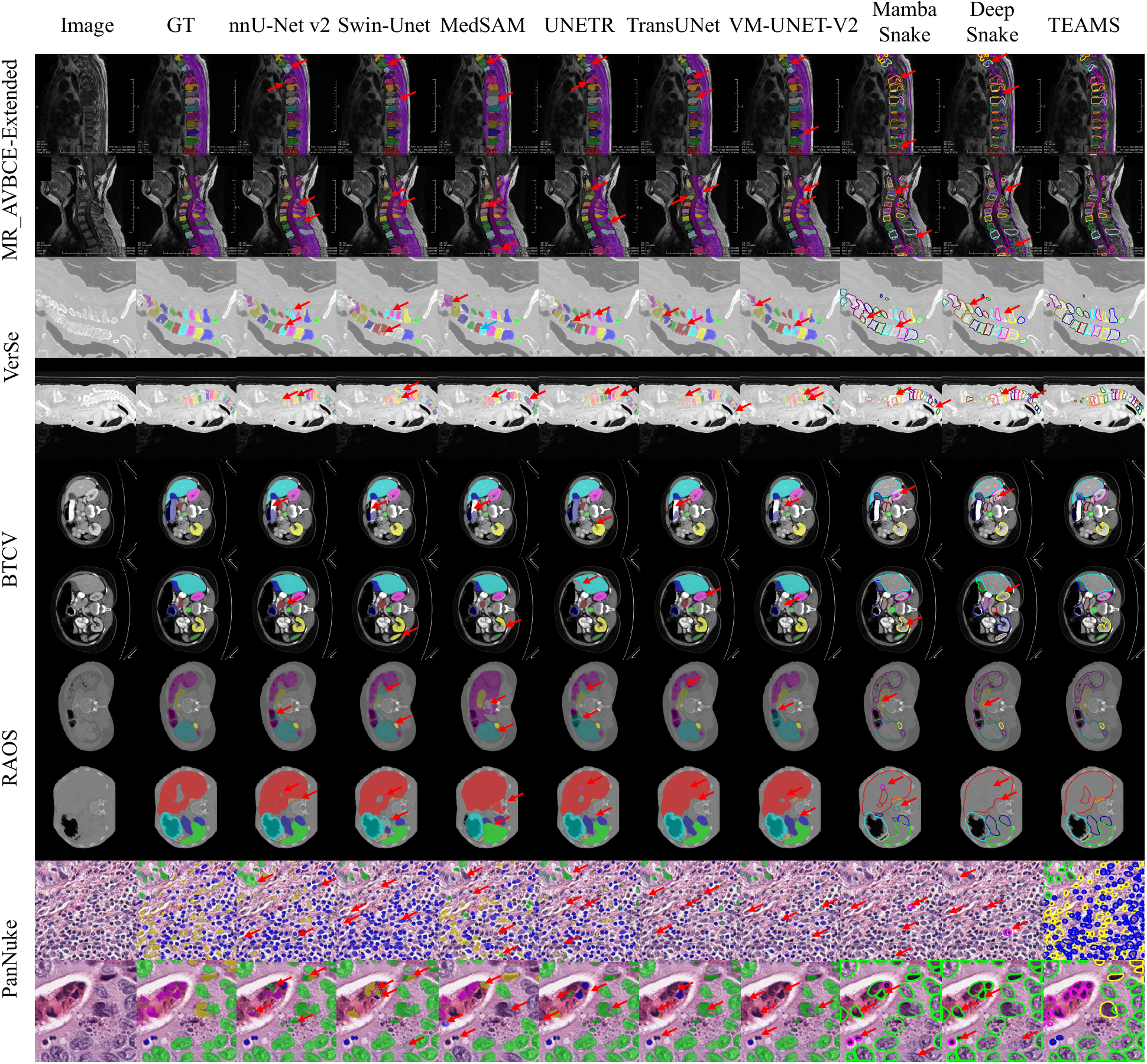}
\caption{TEAMS achieves high instance segmentation performance on five challenging datasets of different imaging modalities and target organs (Column 11). It also outperforms state-of-the-art semantic segmentation (Columns 3$\sim$8) and deep snake (Columns 9/10) methods.}
\label{fig:tcdhs_workflow}
\label{fig8}  %
\vspace{-1em}
\end{figure*}

\section{Experiments and {Discussions}}

\subsection{Datasets, Implementations, and Evaluation Metrics}
\label{Datasets}
\noindent \textbf{4.1.1 Datasets {and challenges}}

We evaluate TEAMS on five challenging multi-organ segmentation datasets: MR\_AVBCE-Extended (\cite{data_MRAVBCE}), VerSe (\cite{data_verse}), RAOS (\cite{data-RAOS}), BTCV (\cite{dataset_btcv}), and PanNuke (\cite{data-PanNuke}). These datasets cover diverse clinical scenarios such as imaging modalities and target organs (spine MRI/CT, abdominal CT, and microscopic nuclei), which are summarized in Table~\ref{tab:datasets_overview_compact}. Representative complex cases are chosen to reflect the challenges and demonstrate the robustness of TEAMS in Fig. \ref{fig8}: 

(1) Spinal MRI/CT exhibits inter-class similarity (different vertebrae bodies and discs in Rows 1$\sim$4), intra-class variation due to occlusion and pathology (the vertebrae and spinal cords can be connected or separated in Rows 1/2, and the pathological morphology changes in T8 of Row 1 and C3 of Row 2), blurred boundaries (the discs in Rows 1/2 and vertebral bodies in Rows 3/4), and densely packed small organs (Row 4). 

(2) Abdominal CT presents complex boundaries (the tortuous left-upper boundary of the left kidney in Row 5/6), blurred boundaries (the low contrast images in Rows 7/8), strong interference (the intensity changes in stomachs/intestines caused by their contents in Rows 5/6/8), and small (and sometimes rare) organs in crowded regions (the colon and adrenal gland in Row 8).

(3) Microscopy nuclei suffer from dense small targets (the number of nuclei is extremely large in Row 9) and similar appearances of different nuclei (the shapes of different classes are difficult to distinguish in Rows 9/10).

\begin{table}[t]
\centering
\footnotesize
\setlength{\tabcolsep}{5pt}
\renewcommand{\arraystretch}{1.2}
\caption{Overview of the five evaluation datasets.}  %
\label{tab:datasets_overview_compact}
\begin{tabular}{l l l c}
\toprule
\textbf{\makecell[l]{Dataset ID/ \\ Name}} & \textbf{\makecell[l]{Modality / \\ Target organs}} & \textbf{Dataset size} & \textbf{\makecell[c]{Target class \\ numbers}} \\ %
\midrule
\addlinespace
\makecell[l]{1. MR\_AVBCE-\\Extended} & MRI / Spine & \makecell[l]{$\sim$480 scans, \\ $\sim$1233 slices} & 50 \\
\addlinespace

2. VerSe & CT / Spine  & \makecell[l]{$\sim$374 scans, \\ $\sim$8000 slices} & 26 \\
\addlinespace

3. BTCV & CT / Abdomen & \makecell[l]{30 scans, \\ $\sim$900 slices}        & 8  \\
\addlinespace

4. RAOS & CT / Abdomen & \makecell[l]{$\sim$317 scans, \\ $\sim$7500 slices} & 19 \\
\addlinespace

5. PanNuke & \makecell[l]{Microscopy / \\ Nuclei} & \makecell[l]{$\sim$7900 \\ image patches} & 5  \\

\bottomrule
\end{tabular}
\end{table}

\noindent \textbf{4.1.2 Input images, ground truth annotations, and implementation details}

The input scans are padded to a square shape and resized to 512$\times$512 for Datasets 1$\sim$4; and 256$\times$256 for Dataset 5. We follow the conference paper (\cite{unified}) for point sampling to obtain the target contours. For training, the Adam optimizer is employed with an initial learning rate of $1 \times 10^{-4}$ and a weight decay of $1 \times 10^{-6}$, which are not fine-tuned in our experiments. The batch size is set as $B=8$. All experiments run on two NVIDIA RTX 4090 GPUs.

\noindent \textbf{4.1.3 Dataset splits}

For Dataset 1, we adopt standard five-fold cross-validation at the scan level. For the other datasets, we follow their official protocols: For Dataset 2, VerSe-2019 uses 80 training scans and 40/40 public/private test scans, while VerSe-2020 uses 113 training scans and 103/103 public/private test scans (\cite{data_verse}). For Dataset 3, we follow the 3:2 train/test split (18/12 patients) (\cite{dataset_btcv}). For Dataset 4, we follow the split of 220/67 patients for train/test (\cite{data-RAOS}). For Dataset 5, we follow the three-fold cross-validation in (\cite{data-PanNuke}). 

{For Datasets 1 and 5, we conduct statistical analyses across different folds. For Datasets 2$\sim$4 with official fixed train/test splits, we keep the benchmark splits unchanged and repeat training with five different random seeds for statistical analyses. To ensure a fair comparison, all baselines are retrained under the same split and training/evaluation protocol.}

\noindent \textbf{4.1.4 Evaluation metrics}

In Datasets 1$\sim$4, we evaluate TEAMS using mean Intersection over Union (mIoU), mean Dice Similarity Coefficient (mDice) (\cite{metric-Dice}), and mean Boundary F (mBF) (\cite{boundf}). The first two metrics are commonly used for assessing segmentation accuracy, while the third indicates target boundary fidelity; we refer readers to (\cite{data_MRAVBCE}) for more details about these metrics. For Dataset 5, we follow the dataset’s official evaluation protocol and adopt multi-class Panoptic Quality (mPQ) as the evaluation metric (\cite{data-PanNuke}), which reflects consistency between detection and segmentation.

\noindent {\textbf{4.1.5 Dataset extension from the conference version}

MR\_AVBCE-Extended dataset substantially extends the MR\_AVBCE dataset in the conference version (\cite{unified}) and significantly increases the dataset difficulty. MR\_AVBCE-Extended contains 1,233 images with more diverse fields of view (FoV), including cervical, thoracolumbar and lumbosacral scans, compared with 600 images from lower-abdominal scans in MR\_AVBCE: In MR\_AVBCE, the vertebral bodies range from S to T9 and the intervertebral discs range from L5/S to T9/T10; however, in MR\_AVBCE-Extended, the vertebral bodies range from S to C1 and the intervertebral discs range from L5/S to C2/C3, increasing the number of target categories to 50. This aggravates the inter-class similarity and challenges discriminative feature learning between similar adjacent structures. Moreover, with this broader anatomical coverage, slight shifts in the scanned spinal range become more common across cases (e.g., from L4–T6 in one case to L5–T7 in another), making existing methods more prone to confusing adjacent classes with similar appearances. MR\_AVBCE-Extended also aggravates intra-class variation by more comprehensively covering cases with occlusion, pathology, and blurred boundaries, making it more challenging than MR\_AVBCE. 
}

\begin{table*}[t]
    \centering
    \caption{Superiority of TEAMS over 14 SOTA methods on five datasets. {Results are reported as mean ± standard deviation.} The best results are highlighted in \textbf{bold} {(this annotation holds for the subsequent tables)}, and the second-best results are underlined. {Statistical significance is marked with * when p < 0.05 compared with the strongest competing method under {two-sided paired t-tests} across folds or repeated runs.}}
    \label{tab:benchmark_no_hd}
    
    \def\tabletwoheaderstretch{1.05}
    \def\tabletworowstretch{2.2}
    \def\tabletwocellstretch{0.90}
    \renewcommand{\arraystretch}{\tabletwoheaderstretch}
    \setlength{\tabcolsep}{2.5pt}
    \renewcommand{\cellset}{\renewcommand{\arraystretch}{\tabletwocellstretch}}
    \makeatletter
    \def\tabletwosetrowstretch#1{%
        \renewcommand{\arraystretch}{#1}%
        \global\setbox\@arstrutbox\hbox{%
            \vrule \@height\arraystretch\ht\strutbox
                   \@depth\arraystretch\dp\strutbox
                   \@width\z@}%
    }
    \makeatother
    \setlength{\aboverulesep}{0.45ex}
    \setlength{\belowrulesep}{0.45ex}
    
    \resizebox{\textwidth}{!}{%
        \begin{tabular}{l | ccc | ccc | ccc | ccc | ccc | ccc | ccc | c}
            \toprule
            \multirow{2}{*}{\textbf{Method}} & 
            \multicolumn{3}{c|}{%
            \makecell[c]{\textbf{MR\_AVBCE-}\\[-1pt]\textbf{Extended}}%
            } &
            \multicolumn{3}{c|}{\textbf{VerSe19(Pu)}} & 
            \multicolumn{3}{c|}{\textbf{VerSe19(Pr)}} & 
            \multicolumn{3}{c|}{\textbf{VerSe20(Pu)}} & 
            \multicolumn{3}{c|}{\textbf{VerSe20(Pr)}} & 
            \multicolumn{3}{c|}{\textbf{BTCV}} &  %
            \multicolumn{3}{c|}{\textbf{RAOS}} &  %
            \textbf{PanNuke} \\ 
            
            \cmidrule(lr){2-4} \cmidrule(lr){5-7} \cmidrule(lr){8-10} \cmidrule(lr){11-13} 
            \cmidrule(lr){14-16} \cmidrule(lr){17-19} \cmidrule(lr){20-22} \cmidrule(lr){23-23}
            
            & mIoU & {mDice} & mBF & mIoU & {mDice} & mBF & mIoU & {mDice} & mBF & mIoU & {mDice} & mBF 
            & mIoU & {mDice} & mBF & mIoU & {mDice} & mBF & mIoU & {mDice} & mBF & mPQ \\
            \midrule
            \noalign{\tabletwosetrowstretch{\tabletworowstretch}}

            \rowcolor{gray!15}
            \rowcolors{1}{gray!15}{gray!15}   %
            \textbf{TEAMS}
            & \makecell{\textbf{69.2$^{*}$}\\$\pm$0.5} & \makecell{\textbf{77.4$^{*}$}\\$\pm$0.4} & \makecell{\textbf{69.6$^{*}$}\\$\pm$0.6}
            & \makecell{\textbf{86.5$^{*}$}\\$\pm$0.4} & \makecell{\textbf{92.5$^{*}$}\\$\pm$0.2} & \makecell{\textbf{82.2$^{*}$}\\$\pm$0.4}
            & \makecell{\textbf{85.9$^{*}$}\\$\pm$0.2} & \makecell{\textbf{91.6}\\$\pm$0.4} & \makecell{\textbf{81.5$^{*}$}\\$\pm$0.4}
            & \makecell{\textbf{87.0$^{*}$}\\$\pm$0.6} & \makecell{\textbf{92.9$^{*}$}\\$\pm$0.3} & \makecell{\textbf{80.7$^{*}$}\\$\pm$0.5}
            & \makecell{\textbf{85.8}$^{*}$\\$\pm$0.4} & \makecell{\textbf{91.8}$^{*}$\\$\pm$0.2} & \makecell{\textbf{80.9$^{*}$}\\$\pm$0.3}
            & \makecell{\textbf{85.3}\\$\pm$0.2} & \makecell{\textbf{91.7$^{*}$}\\$\pm$0.1} & \makecell{\textbf{79.8$^{*}$}\\$\pm$0.4}
            & \makecell{\textbf{79.8$^{*}$}\\$\pm$0.3} & \makecell{\textbf{87.9}$^{*}$\\$\pm$0.4} & \makecell{\textbf{75.9$^{*}$}\\$\pm$0.7}
            & \makecell{\underline{41.1}\\$\pm$0.8} \\

            \midrule

            \makecell[l]{nnU-Net v2 \\ \scriptsize (\cite{compare-nnunet})} 
            & \makecell{62.7\\$\pm$1.1} & \makecell{70.1\\$\pm$1.0} & \makecell{61.8\\$\pm$1.3}
            & \makecell{82.2\\$\pm$0.5} & \makecell{89.7\\$\pm$0.3} & \makecell{76.8\\$\pm$0.6}
            & \makecell{81.0\\$\pm$0.3} & \makecell{88.8\\$\pm$0.4} & \makecell{75.6\\$\pm$0.8}
            & \makecell{80.2\\$\pm$0.4} & \makecell{88.1\\$\pm$0.3} & \makecell{73.9\\$\pm$0.8}
            & \makecell{81.4\\$\pm$0.4} & \makecell{89.5\\$\pm$0.3} & \makecell{75.4\\$\pm$0.8}
            & \makecell{81.9\\$\pm$0.6} & \makecell{86.0\\$\pm$0.9} & \makecell{71.4\\$\pm$0.7}
            & \makecell{73.8\\$\pm$0.2} & \makecell{85.7\\$\pm$0.3} & \makecell{68.6\\$\pm$0.8}
            & \makecell{34.5\\$\pm$1.1} \\

            \makecell[l]{UNETR \\ \scriptsize (\cite{unetr})} 
            & \makecell{56.1\\$\pm$1.1} & \makecell{67.3\\$\pm$1.1} & \makecell{57.7\\$\pm$1.2}
            & \makecell{78.6\\$\pm$0.5} & \makecell{85.3\\$\pm$0.6} & \makecell{71.6\\$\pm$0.9}
            & \makecell{79.7\\$\pm$0.7} & \makecell{85.5\\$\pm$0.6} & \makecell{71.3\\$\pm$0.5}
            & \makecell{78.0\\$\pm$0.6} & \makecell{85.3\\$\pm$0.5} & \makecell{71.8\\$\pm$0.7}
            & \makecell{79.2\\$\pm$0.5} & \makecell{86.4\\$\pm$0.4} & \makecell{72.5\\$\pm$0.9}
            & \makecell{81.0\\$\pm$0.6} & \makecell{84.1\\$\pm$0.7} & \makecell{69.6\\$\pm$1.1}
            & \makecell{74.1\\$\pm$0.7} & \makecell{85.9\\$\pm$0.8} & \makecell{68.1\\$\pm$1.0}
            & \makecell{31.4\\$\pm$1.3} \\

            \makecell[l]{TransUNet \\ \scriptsize (\cite{2021transunet})} 
            & \makecell{55.8\\$\pm$1.0} & \makecell{66.7\\$\pm$1.0} & \makecell{56.4\\$\pm$1.3}
            & \makecell{78.5\\$\pm$0.7} & \makecell{84.5\\$\pm$0.7} & \makecell{70.1\\$\pm$0.9}
            & \makecell{77.8\\$\pm$0.6} & \makecell{84.2\\$\pm$0.6} & \makecell{69.5\\$\pm$0.7}
            & \makecell{78.4\\$\pm$0.6} & \makecell{85.0\\$\pm$0.7} & \makecell{71.2\\$\pm$0.7}
            & \makecell{76.8\\$\pm$0.6} & \makecell{83.9\\$\pm$0.6} & \makecell{69.1\\$\pm$0.9}
            & \makecell{79.0\\$\pm$0.9} & \makecell{85.3\\$\pm$0.9} & \makecell{70.2\\$\pm$1.0}
            & \makecell{72.3\\$\pm$0.6} & \makecell{84.1\\$\pm$0.6} & \makecell{67.2\\$\pm$1.1}
            & \makecell{35.8\\$\pm$1.2} \\

            \makecell[l]{Swin-Unet \\ \scriptsize (\cite{cao2021swinunet})} 
            & \makecell{56.4\\$\pm$1.0} & \makecell{67.0\\$\pm$0.9} & \makecell{57.3\\$\pm$1.4}
            & \makecell{80.8\\$\pm$0.6} & \makecell{87.3\\$\pm$0.3} & \makecell{73.4\\$\pm$1.1}
            & \makecell{80.4\\$\pm$0.6} & \makecell{85.7\\$\pm$0.5} & \makecell{72.6\\$\pm$0.7}
            & \makecell{79.5\\$\pm$0.7} & \makecell{86.6\\$\pm$0.6} & \makecell{72.9\\$\pm$0.5}
            & \makecell{79.8\\$\pm$0.6} & \makecell{85.1\\$\pm$0.5} & \makecell{71.7\\$\pm$0.8}
            & \makecell{80.8\\$\pm$0.6} & \makecell{83.9\\$\pm$0.7} & \makecell{67.8\\$\pm$0.9}
            & \makecell{73.0\\$\pm$0.8} & \makecell{83.6\\$\pm$0.8} & \makecell{66.7\\$\pm$0.9}
            & \makecell{31.3\\$\pm$1.2} \\

            \makecell[l]{VM-UNET-V2 \\ \scriptsize (\cite{vm_net-v2})} 
            & \makecell{57.8\\$\pm$1.3} & \makecell{67.3\\$\pm$1.2} & \makecell{58.6\\$\pm$1.5}
            & \makecell{80.2\\$\pm$0.6} & \makecell{86.4\\$\pm$0.6} & \makecell{72.7\\$\pm$0.9}
            & \makecell{81.1\\$\pm$0.8} & \makecell{86.1\\$\pm$0.4} & \makecell{73.8\\$\pm$0.7}
            & \makecell{79.2\\$\pm$0.5} & \makecell{86.5\\$\pm$0.7} & \makecell{72.1\\$\pm$0.9}
            & \makecell{80.2\\$\pm$0.6} & \makecell{87.1\\$\pm$0.4} & \makecell{73.0\\$\pm$0.8}
            & \makecell{80.1\\$\pm$0.9} & \makecell{85.2\\$\pm$0.6} & \makecell{70.7\\$\pm$0.8}
            & \makecell{73.3\\$\pm$0.8} & \makecell{85.0\\$\pm$0.6} & \makecell{68.3\\$\pm$1.1}
            & \makecell{30.9\\$\pm$1.5} \\

            \makecell[l]{MedSAM \\ \scriptsize (\cite{medsam})} 
            & \makecell{45.9\\$\pm$1.3} & \makecell{50.7\\$\pm$1.2} & \makecell{41.1\\$\pm$1.4}
            & \makecell{53.6\\$\pm$1.0} & \makecell{61.8\\$\pm$0.9} & \makecell{49.3\\$\pm$1.2}
            & \makecell{52.7\\$\pm$0.8} & \makecell{60.7\\$\pm$0.7} & \makecell{48.2\\$\pm$1.0}
            & \makecell{50.9\\$\pm$0.8} & \makecell{58.9\\$\pm$0.8} & \makecell{46.5\\$\pm$1.0}
            & \makecell{51.2\\$\pm$1.0} & \makecell{59.5\\$\pm$1.2} & \makecell{47.7\\$\pm$1.0}
            & \makecell{68.4\\$\pm$1.2} & \makecell{75.3\\$\pm$0.7} & \makecell{62.6\\$\pm$1.1}
            & \makecell{61.7\\$\pm$0.9} & \makecell{67.8\\$\pm$1.2} & \makecell{53.4\\$\pm$1.2}
            & \makecell{26.4\\$\pm$1.6} \\

            \makecell[l]{SIIL \\ \scriptsize (\cite{siil})} 
            & \makecell{62.4\\$\pm$1.1} & \makecell{71.2\\$\pm$0.6} & \makecell{61.3\\$\pm$1.3}
            & \makecell{84.3\\$\pm$0.3} & \makecell{\underline{90.5}\\$\pm$0.4} & \makecell{78.4\\$\pm$0.5}
            & \makecell{\underline{84.8}\\$\pm$0.6} & \makecell{\underline{90.9}\\$\pm$0.3} & \makecell{78.1\\$\pm$0.4}
            & \makecell{\underline{85.5}\\$\pm$0.4} & \makecell{90.0\\$\pm$0.3} & \makecell{77.6\\$\pm$0.5}
            & \makecell{85.0\\$\pm$0.3} & \makecell{\underline{91.2}\\$\pm$0.5} & \makecell{\underline{78.7}\\$\pm$0.7}
            & \makecell{83.7\\$\pm$0.8} & \makecell{86.5\\$\pm$0.4} & \makecell{70.1\\$\pm$0.7}
            & \makecell{73.4\\$\pm$0.7} & \makecell{85.8\\$\pm$0.5} & \makecell{70.3\\$\pm$0.9}
            & \makecell{35.1\\$\pm$1.2} \\

            \makecell[l]{OWT \\ \scriptsize (\cite{owt})} 
            & \makecell{58.4\\$\pm$0.9} & \makecell{69.1\\$\pm$0.9} & \makecell{59.3\\$\pm$1.0}
            & \makecell{81.1\\$\pm$0.6} & \makecell{87.6\\$\pm$0.3} & \makecell{73.5\\$\pm$0.6}
            & \makecell{80.0\\$\pm$0.3} & \makecell{86.3\\$\pm$0.5} & \makecell{72.6\\$\pm$0.4}
            & \makecell{80.4\\$\pm$0.6} & \makecell{86.8\\$\pm$0.4} & \makecell{74.0\\$\pm$0.4}
            & \makecell{79.1\\$\pm$0.5} & \makecell{86.6\\$\pm$0.9} & \makecell{72.9\\$\pm$0.6}
            & \makecell{79.3\\$\pm$0.5} & \makecell{85.4\\$\pm$0.7} & \makecell{66.0\\$\pm$1.0}
            & \makecell{\underline{78.4}\\$\pm$0.7} & \makecell{\underline{87.3}\\$\pm$0.4} & \makecell{73.3\\$\pm$0.6}
            & \makecell{36.2\\$\pm$1.3} \\

            \makecell[l]{TDFormer \\ \scriptsize (\cite{TDformer})} 
            & \makecell{57.6\\$\pm$1.3} & \makecell{67.3\\$\pm$0.8} & \makecell{57.8\\$\pm$1.3}
            & \makecell{82.3\\$\pm$0.6} & \makecell{87.1\\$\pm$0.4} & \makecell{75.7\\$\pm$0.5}
            & \makecell{80.2\\$\pm$0.2} & \makecell{86.9\\$\pm$0.4} & \makecell{72.4\\$\pm$0.7}
            & \makecell{80.7\\$\pm$0.6} & \makecell{87.3\\$\pm$0.6} & \makecell{73.8\\$\pm$0.5}
            & \makecell{79.5\\$\pm$0.6} & \makecell{87.4\\$\pm$0.3} & \makecell{72.6\\$\pm$0.5}
            & \makecell{\underline{84.9}\\$\pm$0.8} & \makecell{\underline{90.1}\\$\pm$0.4} & \makecell{\underline{76.2}\\$\pm$0.7}
            & \makecell{74.3\\$\pm$0.5} & \makecell{85.4\\$\pm$0.7} & \makecell{70.8\\$\pm$0.6}
            & \makecell{35.7\\$\pm$1.2} \\

            \makecell[l]{CellViT \\ \scriptsize (\cite{cellvit})} 
            & \makecell{59.6\\$\pm$0.9} & \makecell{69.3\\$\pm$0.9} & \makecell{60.2\\$\pm$1.2}
            & \makecell{80.0\\$\pm$0.3} & \makecell{87.2\\$\pm$0.3} & \makecell{74.9\\$\pm$0.8}
            & \makecell{78.9\\$\pm$0.5} & \makecell{86.2\\$\pm$0.8} & \makecell{72.8\\$\pm$0.7}
            & \makecell{80.2\\$\pm$0.3} & \makecell{87.6\\$\pm$0.5} & \makecell{74.0\\$\pm$0.8}
            & \makecell{79.6\\$\pm$0.4} & \makecell{86.0\\$\pm$0.3} & \makecell{72.2\\$\pm$0.8}
            & \makecell{80.5\\$\pm$0.2} & \makecell{85.7\\$\pm$0.6} & \makecell{71.3\\$\pm$1.1}
            & \makecell{72.2\\$\pm$0.5} & \makecell{84.1\\$\pm$0.7} & \makecell{66.4\\$\pm$0.8}
            & \makecell{\textbf{43.1}\\$\pm$0.9} \\

            \midrule

            \makecell[l]{Deep Snake \\ \scriptsize (\cite{deep_snake})} 
            & \makecell{56.3\\$\pm$1.2} & \makecell{66.4\\$\pm$1.0} & \makecell{56.7\\$\pm$1.3}
            & \makecell{79.4\\$\pm$0.8} & \makecell{88.0\\$\pm$0.5} & \makecell{75.7\\$\pm$0.6}
            & \makecell{78.6\\$\pm$0.4} & \makecell{86.3\\$\pm$0.9} & \makecell{73.8\\$\pm$0.5}
            & \makecell{78.9\\$\pm$0.4} & \makecell{86.2\\$\pm$0.5} & \makecell{74.2\\$\pm$0.9}
            & \makecell{77.5\\$\pm$0.5} & \makecell{85.7\\$\pm$0.4} & \makecell{72.5\\$\pm$0.6}
            & \makecell{80.4\\$\pm$0.8} & \makecell{85.8\\$\pm$0.9} & \makecell{72.4\\$\pm$1.1}
            & \makecell{72.4\\$\pm$0.8} & \makecell{82.9\\$\pm$0.5} & \makecell{69.3\\$\pm$0.9}
            & \makecell{32.8\\$\pm$1.3} \\

            \makecell[l]{Mamba Snake \\ \scriptsize (\cite{unified})} 
            & \makecell{\underline{63.2}\\$\pm$0.9} & \makecell{71.3\\$\pm$1.1} & \makecell{63.6\\$\pm$1.3}
            & \makecell{\underline{84.4}\\$\pm$0.5} & \makecell{89.6\\$\pm$0.3} & \makecell{\underline{78.8}\\$\pm$0.6}
            & \makecell{84.0\\$\pm$0.3} & \makecell{89.5\\$\pm$0.3} & \makecell{78.1\\$\pm$0.7}
            & \makecell{84.2\\$\pm$0.3} & \makecell{89.6\\$\pm$0.2} & \makecell{77.4\\$\pm$0.4}
            & \makecell{\underline{85.1}\\$\pm$0.6} & \makecell{90.8\\$\pm$0.5} & \makecell{78.4\\$\pm$0.3}
            & \makecell{83.9\\$\pm$0.7} & \makecell{89.6\\$\pm$0.6} & \makecell{75.2\\$\pm$0.9}
            & \makecell{77.3\\$\pm$0.6} & \makecell{86.4\\$\pm$1.1} & \makecell{73.9\\$\pm$0.9}
            & \makecell{35.5\\$\pm$1.3} \\

            \makecell[l]{SAMSnake \\ \scriptsize (\cite{SAMSnake})} 
            & \makecell{61.2\\$\pm$0.8} & \makecell{\underline{72.4}\\$\pm$1.0} & \makecell{\underline{63.8}\\$\pm$1.2}
            & \makecell{84.2\\$\pm$0.6} & \makecell{88.4\\$\pm$0.4} & \makecell{78.0\\$\pm$0.4}
            & \makecell{83.2\\$\pm$0.3} & \makecell{89.8\\$\pm$0.3} & \makecell{\underline{78.3}\\$\pm$0.6}
            & \makecell{84.6\\$\pm$0.4} & \makecell{89.2\\$\pm$0.5} & \makecell{77.2\\$\pm$0.3}
            & \makecell{84.5\\$\pm$0.3} & \makecell{89.0\\$\pm$0.4} & \makecell{77.8\\$\pm$0.9}
            & \makecell{83.4\\$\pm$0.6} & \makecell{89.4\\$\pm$1.0} & \makecell{75.1\\$\pm$0.8}
            & \makecell{78.1\\$\pm$0.6} & \makecell{87.3\\$\pm$0.7} & \makecell{\underline{74.3}\\$\pm$0.5}
            & \makecell{34.2\\$\pm$1.2} \\

            \makecell[l]{ADMIRE \\ \scriptsize (\cite{data_MRAVBCE})} 
            & \makecell{55.1\\$\pm$1.3} & \makecell{63.6\\$\pm$0.8} & \makecell{54.2\\$\pm$1.2}
            & \makecell{81.5\\$\pm$0.6} & \makecell{85.9\\$\pm$0.3} & \makecell{75.8\\$\pm$0.9}
            & \makecell{81.4\\$\pm$0.4} & \makecell{86.0\\$\pm$0.5} & \makecell{75.2\\$\pm$0.2}
            & \makecell{79.1\\$\pm$0.7} & \makecell{85.6\\$\pm$0.3} & \makecell{73.6\\$\pm$0.8}
            & \makecell{82.6\\$\pm$0.6} & \makecell{86.2\\$\pm$0.7} & \makecell{73.1\\$\pm$0.8}
            & \makecell{76.3\\$\pm$0.9} & \makecell{84.2\\$\pm$0.8} & \makecell{69.5\\$\pm$1.1}
            & \makecell{69.2\\$\pm$0.6} & \makecell{78.9\\$\pm$0.8} & \makecell{66.2\\$\pm$1.1}
            & \makecell{31.6\\$\pm$1.2} \\
            \bottomrule
        \end{tabular}%
    }
\end{table*}

\subsection{Qualitative and Quantitative Performances}

\noindent {\textbf{4.2.1 Reliable overall segmentation performance}}

Row 1 of Table \ref{tab:benchmark_no_hd} and Column 11 of Fig. \ref{fig8} show that TEAMS {achieves} reliable contour quality with accurate category discrimination (Column 11 (TEAMS) VS Column 2 (ground truth)). For instance, TEAMS precisely delineates the T7 and T8 vertebrae (orange and gray) with blurred or irregular boundaries while correctly identifying both connected and separated spinal cords (dark purple) in Rows 1/2. Similarly, TEAMS accurately segments small organs such as the left kidney (yellow), the left adrenal (dark orange), and the colon (light gray) despite the low contrast and crowded layouts in Rows 7/8. The quantitative results in the first row of Table \ref{tab:benchmark_no_hd} further confirm that TEAMS consistently achieves high performance across different imaging modalities and target object morphologies.

\noindent {\textbf{4.2.2 Satisfactory segmentation for individual organs}}

\begin{figure}
    \centering
    \includegraphics[width=0.5\textwidth]{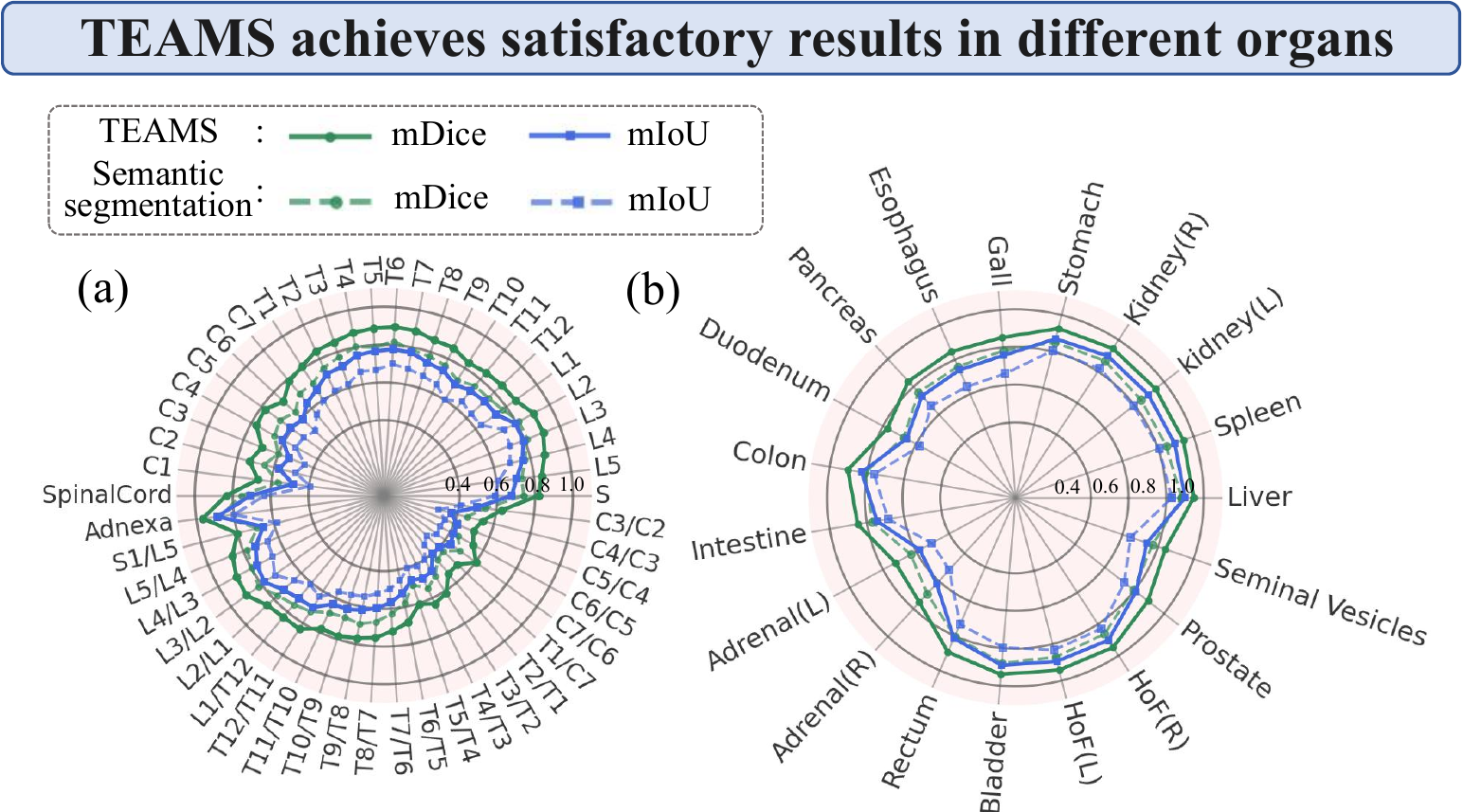}
    \caption{TEAMS consistently achieves strong results for most organ categories on the two most challenging datasets (Datasets 1 and 4). TEAMS (solid lines) outperforms the best semantic segmentation approach (dashed lines) for each organ.}
    \label{fig:mravbce_per_class}
    \vspace{-0.4cm}
\end{figure}

TEAMS also achieves satisfactory results in individual organ categories. 
As shown in Fig. \ref{fig:mravbce_per_class}, TEAMS achieves consistently strong results for most categories, indicating good adaptability to diverse organ boundary morphologies. For a few categories such as the cervical organs in Dataset 1, the performance is relatively lower, probably because these organs rarely occur in the dataset (e.g., the C4/C3 disc appear{s} in only 83 of the 1233 images).  Nevertheless, TEAMS still outperforms competing methods on these categories (the solid (TEAMS) VS the dashed lines (the best compared semantic segmentation approach) in Fig. \ref{fig:mravbce_per_class}(a)). Also, in Dataset 4 where different categories are relatively balanced, TEAMS achieves high mDice and mIoU values for each category (Fig. \ref{fig:mravbce_per_class}(b)).

\subsection{Comparative Experiments}

\noindent {\textbf{4.3.1 Compared methods}}

Extensive comparative experiments with 14 state-of-the-art (SOTA) methods are conducted in the five datasets. To conduct a comprehensive comparison, we include three groups of strong baselines for each dataset: (1) {C}ompetitive universal semantic segmentation models (e.g., nnU-Net v2 (\cite{compare-nnunet}), UNETR (\cite{unetr}), TransUNet (\cite{2021transunet}), Swin-Unet~(\cite{cao2021swinunet}), VM-UNET-V2 (\cite{vm_net-v2}), and MedSAM (\cite{medsam})). (2) Representative deep snake methods (e.g., Deep Snake (\cite{deep_snake}), Mamba Snake (our conference version in \cite{unified}), SAMSnake (\cite{SAMSnake}), and ADMIRE (\cite{data_MRAVBCE})). (3) Recent dataset-specific approaches with publicly reported strong results (e.g., SIIL (\cite{siil}), OWT (\cite{owt}), TDFormer (\cite{TDformer}), and CellViT (\cite{cellvit}) for the four public datasets). 

\noindent {\textbf{4.3.2 Strong performance of TEAMS against SOTA methods}}

Experiments show that TEAMS achieves {performance} comparable to or better than {that of} the 14 SOTA methods in the five datasets. Row 1 of Table \ref{tab:benchmark_no_hd} shows that TEAMS leads on Datasets 1$\sim$4. It shows a relative mDice/mBF improvement of {6.9\%/9.1\%} compared with the second-best method in Dataset 1. On Dataset 5, the mPQ ranks second among the competing methods, only slightly lower than the dataset-specific approach on this dataset. Moreover, deep snake's {advantage over semantic segmentation} is demonstrated by comparing their performances against semantic segmentation (Rows 1, 12$\sim$15 VS 2$\sim$11). {We further conduct two-sided paired t-tests between TEAMS and the strongest competing method across the same folds or random seeds. As shown in Table \ref{tab:benchmark_no_hd}, the small standard deviations show that TEAMS achieves stable performance improvements across different folds or random seeds, and the significance tests further indicate that its improvements are not caused by random fluctuations in a single run, with most p-values below 0.05.}

The qualitative results in Fig. \ref{fig8} further support this trend. The arrows in Columns 3$\sim$8 of Fig. \ref{fig8} show that semantic segmentation may yield illogical segmentation in complex cases. For example, they confuse adjacent tissues or yield wrong organ shapes when encountering blurred boundaries and/or densely packed organs (arrows in Rows 1$\sim$8). Also, they could incorrectly segment or miss small targets (the colon and adrenal gland in Row 8 and missing nuclei in Rows 9/10). 
These inferior segmentations are also observed in other articles (\cite{9201093,siil,TDformer}), demonstrating that these illogical results are no accident. In contrast, the deep snake methods yield anatomically reasonable results (Columns 9$\sim$10). With the designed core modules, TEAMS (Column 11) further enhances instance segmentation performance.

\subsection{Ablation Study}

\noindent \textbf{4.4.1 Contribution of each design}

\begin{figure}
    \centering
    \includegraphics[width=\linewidth]{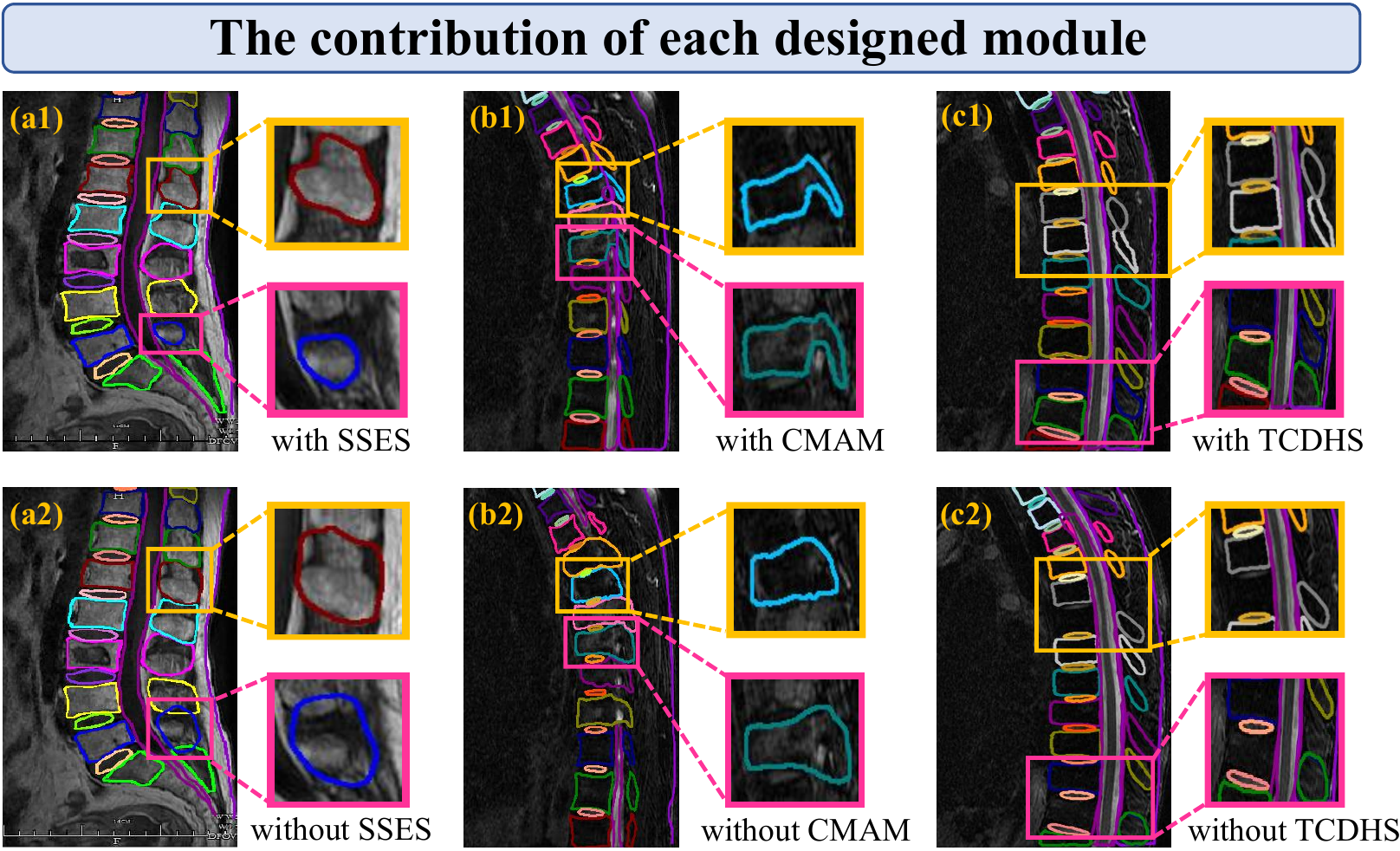}
    \caption{The contributions of each core module. When each core module is enabled, the evolution failures are mitigated (a1 VS a2), the boundary details are better delineated (b1 VS b2), and the wrong detections are avoided (c1 VS c2).}
    \label{fig:core_modules}

\end{figure}

Ablation experiments on Dataset 1 show that the three core designs of TEAMS bring improvements over a strong baseline. We start with the baseline of "YOLOv8 detection head + deep snake evolution" (Row 1 of Table \ref{tab:ablation_study}), and add SSES, CMAM, text prompts in TCDHS, detection consistency feedback in TCDHS, and the full TCDHS in Rows 2$\sim$6 of Table \ref{tab:ablation_study}. As shown in Row 1 in Table \ref{tab:ablation_study}, the baseline achieves a satisfactory performance (ranks in the top half compared with the semantic segmentation methods in Table \ref{tab:benchmark_no_hd}).
With the addition of SSES, CMAM, text prompts, detection consistency feedback, and full TCDHS, the mDice shows relative improvements of 8.0\%, 5.6\%, 5.6\%, 6.1\%, and 7.3\% (Rows 2$\sim$6 of Table \ref{tab:ablation_study}), showing that deep snake methods could benefit from these designs. The qualitative results in Fig. \ref{fig:core_modules}(a)$\sim$(c) further support this conclusion: {SSES guides} the snake to the target contour (Fig. \ref{fig:core_modules}(a1)){, whereas its absence causes} evolution failure (Fig. \ref{fig:core_modules}(a2)). CMAM accurately delineate{s} boundary details in complex regions (Fig. \ref{fig:core_modules}(b1)), {whereas its absence causes} over-smoothing (Fig. \ref{fig:core_modules}(b2)). TCDHS reduces wrong detections (Fig. \ref{fig:core_modules}(c1)) compared with the ablated version (Fig. \ref{fig:core_modules}(c2)).

\begin{table}
\centering
\begin{minipage}[t]{\linewidth}
    \centering
    \captionsetup{type=table, justification=centering, skip=3pt, font=small}
    \captionof{table}{Ablation study demonstrating the effect of SSES, CMAM, text prompts, detection consistency feedback, and full TCDHS.}
    \label{tab:ablation_study}
    \renewcommand{\arraystretch}{1.05}
    \setlength{\tabcolsep}{0.85pt}
    \resizebox{1.0\linewidth}{!}{
        \begin{tabular}{lccc}
        \toprule
        \textbf{Model Configuration} & \textbf{mIoU} & \textbf{mDice} & \textbf{mBF}  \\
        \midrule
        Baseline
        & 58.2
        & 67.5
        & 58.8 \\

        \hspace{0.5em}{+ SSES}
        & 62.2({$\Uparrow$6.9\%})
        & 72.9({$\Uparrow$8.0\%})
        & 62.8({$\Uparrow$6.8\%}) \\

        \hspace{0.5em}{+ CMAM}
        & 59.1({$\Uparrow$1.5\%})
        & 71.3({$\Uparrow$5.6\%})
        & 60.3({$\Uparrow$2.6\%}) \\

        \hspace{0.5em}{+ Text prompts in TCDHS}
        & 60.7({$\Uparrow$4.3\%})
        & 71.3({$\Uparrow$5.6\%})
        & 61.4({$\Uparrow$4.4\%}) \\

        \hspace{0.5em}{+ Consistency feedback in TCDHS}
        & 61.3({$\Uparrow$5.3\%})
        & 71.6({$\Uparrow$6.1\%})
        & 60.8({$\Uparrow$3.4\%}) \\

        \hspace{0.5em}{+ TCDHS}
        & 61.9({$\Uparrow$6.3\%})
        & 72.4({$\Uparrow$7.3\%})
        & 62.5({$\Uparrow$6.3\%}) \\

        \textbf{TEAMS}
        & \textbf{69.2}({\textbf{$\Uparrow$18.9\%}})
        & \textbf{77.4}({\textbf{$\Uparrow$14.7\%}})
        & \textbf{69.6}({\textbf{$\Uparrow$18.4\%}}) \\
        \bottomrule
        \end{tabular}
    }
\end{minipage}
\hfill
\end{table}

\begin{figure}
    \centering
    \includegraphics[width=\linewidth]{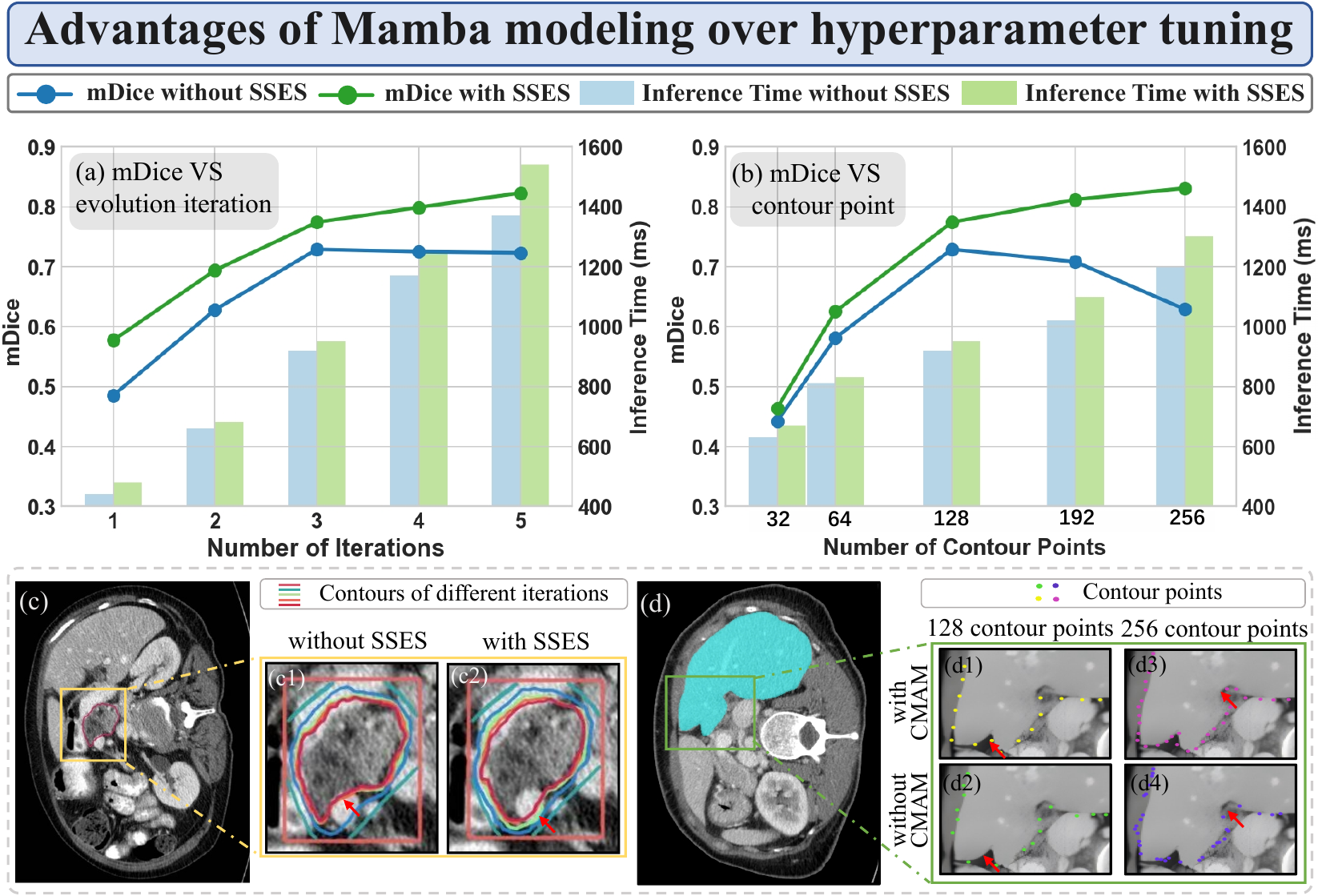}
    \caption{The necessity of Mamba modeling in deep snakes. Fig. \ref{fig:iter_and_points}(a)/(b) show SSES/CMAM (green) better benefits overall performance than hyperparameter tuning (blue). Fig. \ref{fig:iter_and_points}(c) shows SSES (c1) better mitigates evolution failure than increasing iteration numbers (c2). Fig. \ref{fig:iter_and_points}(d) shows CMAM (d1) better delineates fine-grained organ details than increasing contour points (d2). }
    \label{fig:iter_and_points}
    \vspace{-1em}
\end{figure}

{
\noindent \textbf{4.4.2 The benefits of the temporal modeling strategies and exponential decay aggregation in SSES} 

Ablation experiments further validate the benefits of SSES over standard temporal modeling methods. Specifically, two ablated versions replacing SSES with standard temporal feature aggregation and GRU-based recurrent modeling are constructed for comparison. The experimental results are shown in Table \ref{tab:sses_temporal_comparison}, which shows that SSES achieves the best performance in mIoU, mDice, and mBF. The performance improvement over standard temporal feature aggregation is consistent with the mechanism analysis in Section \ref{sec:SSES}, suggesting that SSES benefits snake evolution by leveraging informative trajectory cues and contour point contexts beyond simple weighted fusion. Also, the improvement over GRU-based recurrent modeling shows that the contour morphology-aware structured state space duality serves as a better computational unit than recurrent modeling. Thus, the proposed SSES is effective in improving evolution vector prediction and final segmentation performance.

\begin{table}[t]
\centering
\begin{minipage}[t]{\linewidth}
    \centering

    \captionsetup{
        justification=centering,
        skip=3pt,
        font=small
    }

    \caption{
        SSES achieves superior performance compared with existing
        temporal modeling strategies.
    }
    \label{tab:sses_temporal_comparison}

    {

    \renewcommand{\arraystretch}{1.05}

    {\teamsTableFont
    \setlength{\tabcolsep}{4pt}

    \begin{tabularx}{\linewidth}{
        >{\raggedright\arraybackslash}X
        >{\centering\arraybackslash}c
        >{\centering\arraybackslash}c
        >{\centering\arraybackslash}c
    }

        \toprule

        \textbf{Temporal modeling strategy}
        & \textbf{mIoU}
        & \textbf{mDice}
        & \textbf{mBF}
        \\

        \midrule

        Standard temporal feature aggregation
        & 66.6
        & 73.6
        & 66.5
        \\

        GRU-based recurrent modeling
        & 66.1
        & 72.9
        & 62.9
        \\

        SSES
        & \textbf{69.2}
        & \textbf{77.4}
        & \textbf{69.6}
        \\

        \bottomrule
    \end{tabularx}
    }

    } %

\end{minipage}
\end{table}

Another ablated version using attention-based temporal fusion is compared with exponential decay aggregation, demonstrating that exponential decay aggregation is more reasonable than attention-based temporal fusion in SSES. As reported in Table \ref{tab:exponential_decay_comparison}, the attention-based temporal fusion underperforms exponential decay aggregation while increasing computing cost. Thus, for the iterative convergence process of snake evolution, explicitly assigning larger weights to more recent states is more accurate and efficient than using attention-based temporal fusion to learn temporal weights.

\begin{table}[t]
\centering
\begin{minipage}[t]{\linewidth}
    \centering

    \captionsetup{
        justification=centering,
        skip=3pt,
        font=small
    }

    \caption{
        The superiority of exponential decay aggregation over
        attention-based temporal fusion.
    }
    \label{tab:exponential_decay_comparison}

    {

    \renewcommand{\arraystretch}{1.05}

    {\teamsTableFont
    \setlength{\tabcolsep}{1.5pt}

   \setlength{\tabcolsep}{0.5pt}   
   
\begin{tabular}{@{}
        >{\raggedright\arraybackslash}p{0.235\linewidth}
        *{3}{>{\centering\arraybackslash}p{0.095\linewidth}}
        *{3}{>{\centering\arraybackslash}p{0.15\linewidth}}
        @{}}
    \toprule
    \multicolumn{1}{@{}l}{%
        \raisebox{0.8ex}{%
            \makecell[tl]{%
                \textbf{Temporal modeling}\\
                \textbf{method}%
            }%
        }%
    }
    & \textbf{mIoU}
    & \textbf{mDice}
    & \textbf{mBF}
    & \makecell[c]{\textbf{Trainable}\\\textbf{params (M)}}
    & \makecell[c]{\textbf{Inference}\\\textbf{time (s)}}
    & \makecell[c]{\textbf{GPU mem-}\\\textbf{ory (MB)}}
    \\
    \midrule
    \begin{tabular}[c]{@{}l@{}}
        Attention-based\\
        temporal fusion
    \end{tabular}
    & 63.0 & 73.9 & 63.1 & 17.67 & 1.25 & 3066 \\
    \begin{tabular}[c]{@{}l@{}}
        Exponential decay\\
        aggregation (ours)
    \end{tabular}
    & \textbf{69.2} & \textbf{77.4} & \textbf{69.6}
    & \textbf{15.74} & \textbf{1.17} & \textbf{2852} \\
    \bottomrule
\end{tabular}

    }
    } %

\end{minipage}
\end{table}
}

{
\noindent \textbf{4.4.3 The benefits of local geometric guidance in CMAM}

To further validate the robustness and generalization ability of CMAM, an ablated version that replaces the explicit local geometric guidance with the learned alternative is compared with TEAMS. In the learned alternative, the guidance provided by $\mathbf{L}^*$ is removed so that the structured attention mask is learned solely from data. The results are reported in Table \ref{tab:cmam_learned_alternative}, which shows that our CMAM outperforms the learned alternatives on all five datasets, indicating that the morphology guidance constructed from $\alpha_n$, $\kappa_n$, and $\rho_n$ is beneficial for structured attention mask learning. Since these datasets cover different modalities, organs, and target morphologies, the consistent improvements indicate that the benefit of morphology-guided mask learning is not limited to a specific scenario. In contrast, the learned alternative lacks explicit local morphology guidance and shows weaker performance. These results directly support the robustness and generalization ability of the CMAM design.

\begin{table}[t]
\centering
\begin{minipage}[t]{\linewidth}
    \centering

    \captionsetup{
        justification=centering,
        skip=3pt,
        font=small
    }

    \caption{
        The morphology-guided learnable structured attention mask modeling
        in CMAM is better than the learned alternative.
    }
    \label{tab:cmam_learned_alternative}

    {

    \renewcommand{\arraystretch}{1.14}

    {\scriptsize
    \setlength{\tabcolsep}{1.5pt}

    \begin{tabularx}{\linewidth}{@{}
        >{\raggedright\arraybackslash}p{0.22\linewidth}
        @{\hspace{4pt}}
        >{\raggedright\arraybackslash}X
        @{\hspace{3pt}}
        *{4}{>{\centering\arraybackslash}p{0.11\linewidth}}
        @{}}

        \toprule

        \textbf{Dataset}
        & \textbf{Method}
        & \textbf{mIoU}
        & \textbf{mDice}
        & \textbf{mBF}
        & \textbf{mPQ}
        \\

        \midrule

        \multirow[c]{2}{=}{\raggedright MR\_AVBCE-\\Extended}
        & Learned alternative
        & 61.9
        & 69.9
        & 60.4
        & -- \\

        & Our CMAM
        & \textbf{69.2}
        & \textbf{77.4}
        & \textbf{69.6}
        & -- \\

        \addlinespace[2pt]

        \multirow[c]{2}{*}{VerSe}
        & Learned alternative
        & 80.4
        & 83.1
        & 71.0
        & -- \\

        & Our CMAM
        & \textbf{87.0}
        & \textbf{92.9}
        & \textbf{80.7}
        & -- \\

        \addlinespace[2pt]

        \multirow[c]{2}{*}{BTCV}
        & Learned alternative
        & 79.7
        & 82.4
        & 70.6
        & -- \\

        & Our CMAM
        & \textbf{85.3}
        & \textbf{91.7}
        & \textbf{79.8}
        & -- \\

        \addlinespace[2pt]

        \multirow[c]{2}{*}{RAOS}
        & Learned alternative
        & 72.3
        & 82.6
        & 66.5
        & -- \\

        & Our CMAM
        & \textbf{79.8}
        & \textbf{87.9}
        & \textbf{75.9}
        & -- \\

        \addlinespace[2pt]

        \multirow[c]{2}{*}{PanNuke}
        & Learned alternative
        & --
        & --
        & --
        & 40.5 \\

        & Our CMAM
        & --
        & --
        & --
        & \textbf{41.1} \\

        \bottomrule
    \end{tabularx}
    }

    } %

\end{minipage}
\end{table}

}

\noindent \textbf{4.4.4 Advantages of Mamba modeling over intuitive hyperparameter tuning}

To further validate the necessity of Mamba modeling in SSES and CMAM, we examine whether hyperparameter tuning could replace these designs. Intuitively, increasing the number of evolution iterations may give the snake more opportunities to reach the target boundary and potentially alleviate evolution failure; increasing the number of contour points may improve contour resolution and thus enhance boundary details. However, {the blue lines in Fig. \ref{fig:iter_and_points}(a)/(b)} show that these strategies cannot achieve the desired result. In contrast, SSES enables better exploitation of additional iterations (green line in Fig. \ref{fig:iter_and_points}(a)), and CMAM enables better utilization of additional points (green line in Fig. \ref{fig:iter_and_points}(b)). As further visualized in Fig. \ref{fig:iter_and_points}(c)/(d), SSES helps the snake to reach the targets while the snake could stagnate when simply {increasing the number of iterations} (c1 VS c2); CMAM correctly evolves the points to delineate fine-grained details while simply adding contour points does not {achieve correct delineation} (the arrows in (d1)/(d3) VS (d2)/(d4)).
Lastly, the extra inference time introduced by adding SSES/CMAM is less than directly adding iterations/contour points (the green bars VS the next blue bars in Fig. \ref{fig:iter_and_points}(a) and (b)).

{
\noindent \textbf{4.4.5 Advantages of Mamba modeling over Transformer-based and convolution-based alternatives}

To further justify the advantage of Mamba-based sequence modeling in snake evolution, ablated versions in which the Mamba operations are replaced with Transformer or convolutional operations are compared with TEAMS. The results are reported in Table \ref{tab:mamba_operator_comparison}, which shows that Mamba snake evolution achieves the best performance. Compared with the {convolution-based alternative}, this improvement is consistent with the analysis in Section \ref{sec:CMAM} that {fixed-size} kernels in CNNs are less flexible for modeling contour neighborhoods with different morphological complexity. The comparison with the {Transformer-based alternative} also shows that Mamba-based selective state space modeling better supports order-aware feature propagation along the contour. Therefore, Mamba is more suitable as the core sequence modeling operator in the context of snake evolution.

\begin{table}[t]
\centering
\begin{minipage}[t]{\linewidth}
    \centering

    \captionsetup{
        justification=centering,
        skip=3pt,
        font=small
    }

    \caption{
        Mamba-based snake evolution outperforms convolution-based
        and Transformer-based alternatives.
    }
    \label{tab:mamba_operator_comparison}

    {

    \renewcommand{\arraystretch}{1.05}

    {\teamsTableFont
    \setlength{\tabcolsep}{4pt}

    \begin{tabularx}{\linewidth}{
        >{\raggedright\arraybackslash}X
        >{\centering\arraybackslash}c
        >{\centering\arraybackslash}c
        >{\centering\arraybackslash}c
    }
        \toprule

        \textbf{Variant}
        & \textbf{mIoU}
        & \textbf{mDice}
        & \textbf{mBF}
        \\

        \midrule

        Convolution-based snake evolution
        & 58.9
        & 66.0
        & 60.9
        \\

        Transformer-based snake evolution
        & 63.5
        & 72.8
        & 64.9
        \\

        Mamba-based snake evolution (ours)
        & \textbf{69.2}
        & \textbf{77.4}
        & \textbf{69.6}
        \\

        \bottomrule
    \end{tabularx}
    }

    } %

\end{minipage}
\end{table}
}

\begin{figure}
    \centering
    \includegraphics[width=\linewidth]{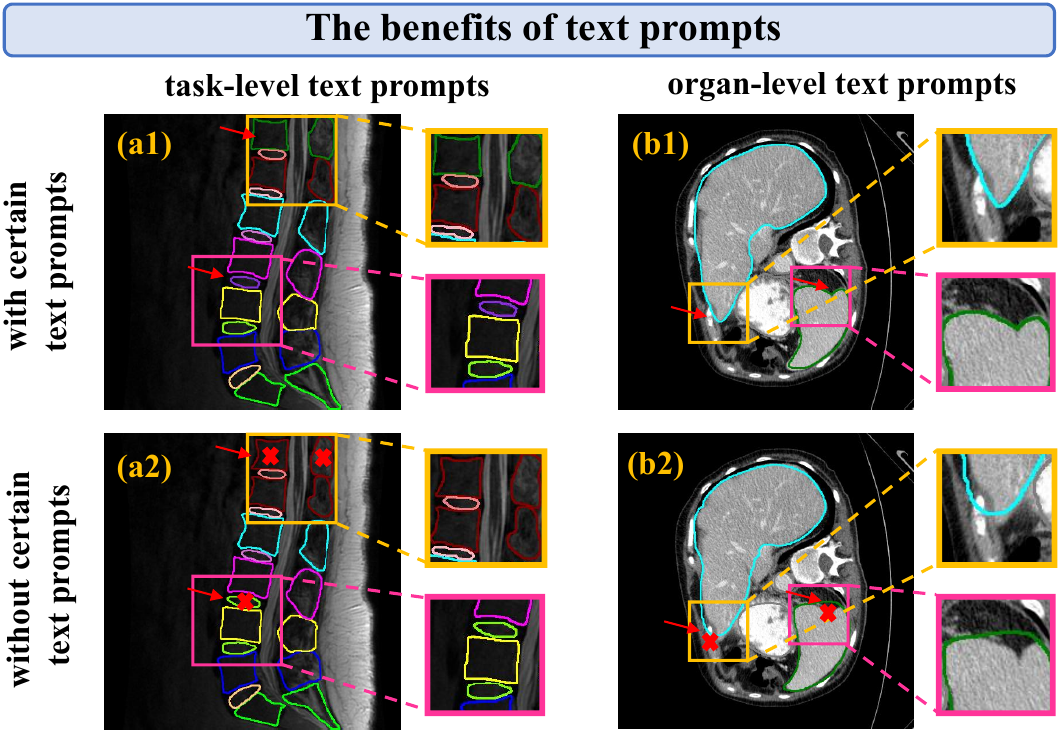}
    \caption{The benefits of the text prompts. The task-level text prompts help to better distinguish different organs (a1 VS a2), and the organ-level text prompts help the snake to better evolve to target organs (b1 VS b2).}
    \label{fig:text prompts effectiveness}
    \vspace{-1em}
\end{figure}

{
\noindent \textbf{4.4.6 The benefits of dual-head consistency feedback}

This mechanism benefits deep snake segmentation across all three evaluation metrics with statistically significant improvements. The mDice, mIoU, and mBF show relative improvements of respectively 6.1\%, 5.3\%, and 3.4\%. These consistent improvements indicate that the dual-head consistency feedback reliably improves segmentation performance. A statistical significance analysis under {five}-fold cross-validation on Dataset 1 further validates the reliability of these improvements. This analysis shows that the p-values of mDice, mIoU, and mBF are 0.012, 0.018, and 0.028, highlighting that these improvements are statistically significant. These improvements demonstrate that the dual-head consistency feedback benefits the deep snake workflow.

The visualizations in Fig. \ref{fig8} also validate that TEAMS is less prone to missing detections. To further analyze this phenomenon, we have examined the confidence scores assigned to the correct classes of all true targets. The proportion of true targets whose confidence scores for the correct classes fall into a low confidence range (e.g., score < 0.1) decreases from 9.7\% (without the dual-head feedback) to 2.3\%  (with the dual-head feedback) on Dataset 3, i.e., much fewer true targets fall into the low confidence range where they are difficult to distinguish from background or false positive proposals and are thus more vulnerable to being filtered out before snake initialization. These results indicate that the detection quality is improved and the risk of “completely missed” targets is effectively mitigated in practice.
}

\noindent \textbf{4.4.7 The benefits of text prompts}

For overall performance, adding the text prompts results in a mDice increase of 5.6\% (Row 4 in Table \ref{tab:ablation_study}). For detailed visualization,  Fig. \ref{fig:text prompts effectiveness}(a) shows that task-level text prompts such as "An MRI spine dataset, including organs such as vertebrae bodies (the vertebrae labels from bottom to up are S, L5, L4, ... C2, C1), discs (are located between the corresponding vertebrae, labels from bottom to top are S/L5, L5/L4, ... C3/C2) ..." provide generic topological ordering of the spinal organs, helping TEAMS to better distinguish different organs (the arrows in Fig. \ref{fig:text prompts effectiveness}(a1)). Without this information, more wrong detections may occur (the arrows in Fig. \ref{fig:text prompts effectiveness}(a2)). Similarly, Fig. \ref{fig:text prompts effectiveness}(b) shows that organ-level text prompts such as "The spleen is a highly variable organ, which may appear as wedge, circle or heart, lateral and posterior to the stomach..." and "The left lateral lobe of liver could taper to a sharp pointed edge ..." help the snake to reach the target contours (the arrows in Fig. \ref{fig:text prompts effectiveness}(b1)). Without the shape information, the snake evolution head may yield sub-optimal segmentation (the arrows in Fig. \ref{fig:text prompts effectiveness}(b2)).

{
\noindent \textbf{4.4.8 Analysis on different types or qualities of prompts}

To further validate the role of textual prompts in TEAMS, ablation studies such as no prompt, random prompt, incorrect prompt, different correct templates, are carried out and compared with TEAMS. The no prompt setting evaluates the performance after removing text input. The incorrect prompt and random prompt settings evaluate the effects of incorrect textual semantics or incorrect text-target correspondence. The different correct templates setting evaluates whether the model depends on a fixed prompt sentence template. The results in Table \ref{tab:vl_prompt_ablation} show that TEAMS achieves the best performance. When the prompts change, we have the following observations: 
\begin{itemize}
    \item Correct textual prompts are consistently beneficial: As long as the textual prompts are correct, the different correct templates consistently outperform the no prompt setting and yield relatively stable results across templates (Rows 4, 5, and 6 VS Row 1 in Table \ref{tab:vl_prompt_ablation}), showing that TEAMS does not rely on a fixed sentence template but can benefit from correct textual prompts to guide detection and snake evolution. 
    \item Incorrect or irrelevant textual prompts weaken the performance: In contrast, the performance decreases when random / incorrect text is used (worse than using no prompt, Rows 2 and 3 VS Row 1 in Table \ref{tab:vl_prompt_ablation}), which indicates that the improvement depends on correct textual semantics and text-target correspondence.
\end{itemize}
These results demonstrate that TEAMS benefits from semantically correct textual prompts across different templates, whereas low-quality or semantically incorrect prompts may degrade performance. This further clarifies the role of textual features in TEAMS: they participate in vision-language feature modeling and guide downstream predictions in the "detection-then-evolution" deep snake workflow.

\begin{table}[t]
\centering
\begin{minipage}[t]{\linewidth}
    \centering

    \captionsetup{
        justification=centering,
        skip=3pt,
        font=small
    }

    \caption{
        Ablation study on different textual prompt settings validates that textual prompts benefit snake segmentation.
    }
    \label{tab:vl_prompt_ablation}

    {

    \renewcommand{\arraystretch}{1.05}

    {\teamsTableFont
    \setlength{\tabcolsep}{4pt}

    \begin{tabularx}{\linewidth}{
        >{\raggedright\arraybackslash}X
        >{\centering\arraybackslash}c
        >{\centering\arraybackslash}c
        >{\centering\arraybackslash}c
    }
        \toprule

        \textbf{Setting}
        & \textbf{mIoU}
        & \textbf{mDice}
        & \textbf{mBF}
        \\

        \midrule

        No prompt
        & 65.8
        & 73.6
        & 64.3
        \\

        Random prompt
        & 63.3
        & 71.9
        & 62.8
        \\

        Incorrect prompt
        & 63.0
        & 71.3
        & 62.1
        \\

        Different correct template 1
        & 68.9
        & \textbf{77.4}
        & 68.8
        \\

        Different correct template 2
        & 68.5
        & 76.7
        & 68.5
        \\

        Correct prompt / TEAMS
        & \textbf{69.2}
        & \textbf{77.4}
        & \textbf{69.6}
        \\

        \bottomrule
    \end{tabularx}
    }

    } %

\end{minipage}
\end{table}
}

{\subsection{Computational Costs and Failure Cases}

\noindent \textbf{4.5.1 Computational costs}

Table \ref{tab:computational_cost} reports the computational cost of TEAMS and the conference version. TEAMS contains 15.74M trainable parameters; it requires 1.17 s inference time and 2852 MB peak GPU memory. Compared with the conference version, most parameters in TEAMS come from the frozen ClinicalBERT text encoder (137.43M), whereas the trainable parameters remain limited (15.74M). Since the ClinicalBERT text features are extracted offline and cached, TEAMS does not need to repeatedly run the text encoder, keeping the inference time moderate (1.17s). Furthermore, TEAMS maintains moderate peak GPU memory consumption (2852 MB) while achieving strong segmentation performance. 

\begin{table}[t]
\centering
\begin{minipage}[t]{\linewidth}
    \centering

    \caption{
        The computational costs of TEAMS and the conference version on MR\_AVBCE-Extended.
    }
    \label{tab:computational_cost}

    {

    \renewcommand{\arraystretch}{1.18}

    \scriptsize
    \setlength{\tabcolsep}{2pt}

    \begin{tabularx}{\linewidth}{@{}
        >{\raggedright\arraybackslash}p{0.235\linewidth}
        >{\centering\arraybackslash}X
        >{\centering\arraybackslash}p{0.145\linewidth}
        >{\centering\arraybackslash}p{0.18\linewidth}
        >{\centering\arraybackslash}p{0.145\linewidth}
        @{}}
        \toprule

        \textbf{Method}
        & \textbf{Params (M)}
        & \textbf{\makecell[c]{Inference\\time (s)}}
        & \textbf{\makecell[c]{Peak GPU\\memory (MB)}}
        & \textbf{\makecell[c]{mDice/\\mBF}}
        \\

        \midrule

        \makecell[l]{Conference\\version}
        & 12.48
        & 0.83
        & 1856
        & 71.3/63.6
        \\

        \addlinespace[2pt]

        \makecell[l]{Journal version\\(TEAMS)}
        & \makecell[c]{15.74 (trainable)\\+ 137.43 (frozen)}
        & 1.17
        & 2852
        & 77.4/69.6
        \\

        \bottomrule
    \end{tabularx}

    } %

\end{minipage}
\end{table}

\vspace{0.4cm}  %
\noindent \textbf{4.5.2 Failure cases}

TEAMS may produce suboptimal results in cases of annotation ambiguity. For example, the solid green lines in Fig. \ref{fig:failure_case} show inconsistent annotations of vessels passing through the liver, where vessels may be either excluded from (Fig. \ref{fig:failure_case}(a1)) or included in (Fig. \ref{fig:failure_case}(a2)) the liver region. Under such annotation inconsistency, the snake evolution could be disturbed, and TEAMS could produce suboptimal segmentation results. This suggests that annotation consistency remains important for TEAMS.

\begin{figure}
    \centering
    \includegraphics[width=\linewidth]{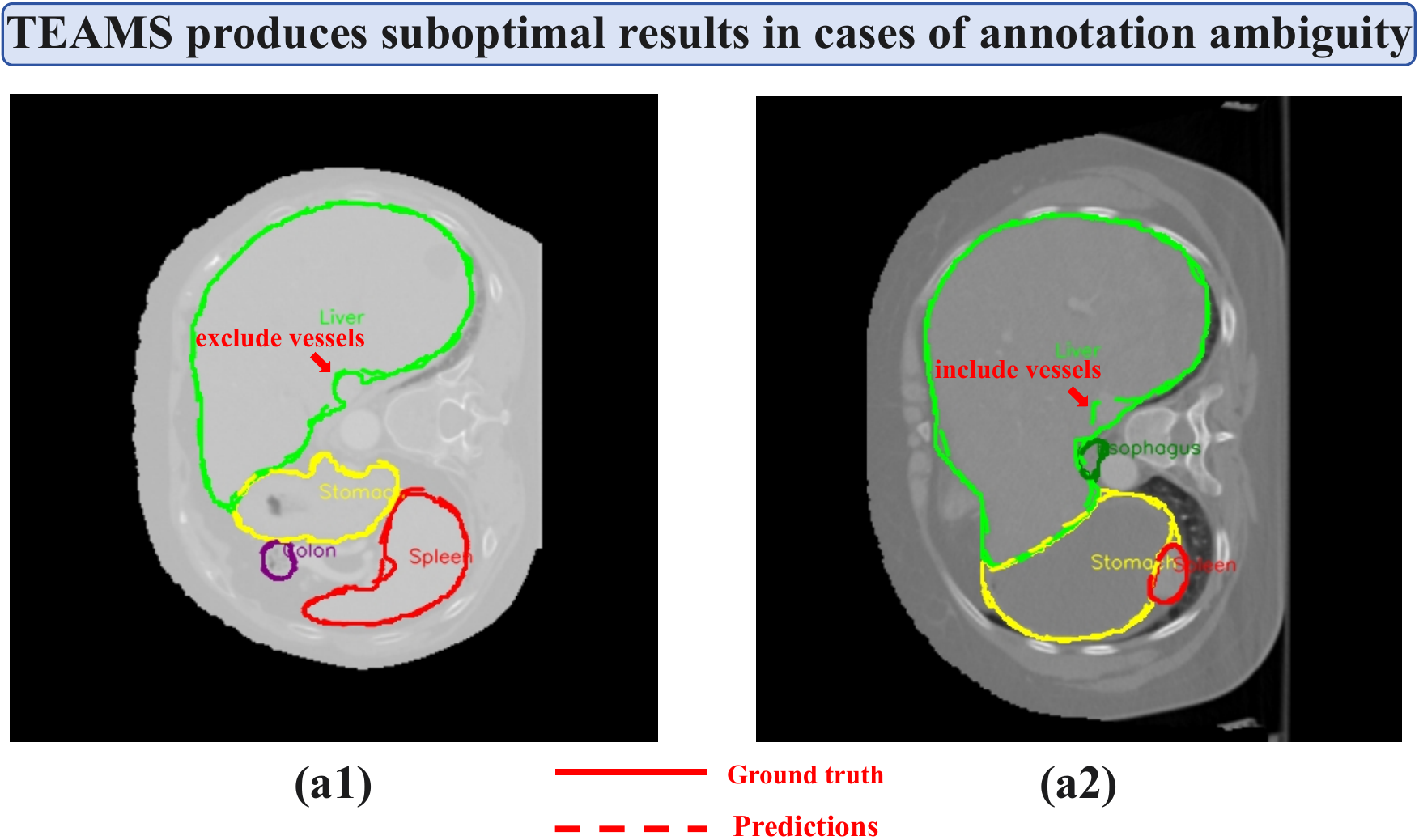}

    \begingroup
    \caption{{
        Representative failure cases of TEAMS under annotation ambiguity. The solid contours denote the ground truth, and the dashed contours denote the evolved snake predictions. Red arrows indicate ambiguous vessel regions near the liver boundary: the vessel is excluded from the liver ground truth in Fig. \ref{fig:failure_case}(a1), but included in the liver ground truth in Fig. \ref{fig:failure_case}(a2). Such inconsistent annotations may lead to suboptimal segmentation results.
    }}
    \label{fig:failure_case}
    \endgroup

    \vspace{-1em}
\end{figure}

}
{\subsection{Potential Clinical Applications and Extensions}}
TEAMS provides object-level contours for multiple targets across diverse imaging modalities, which can support comprehensive multi-organ diagnosis and treatment planning. The explicit organ/lesion boundaries could directly provide diagnostic information (e.g., interface clarity and margin irregularity) for diagnosing spinal tumors (\cite{ross2020diagnostic}), supporting treatment planning in abdominal radiotherapy (\cite{shi2022deep}), and assessing nuclear morphology for grading cancers in histopathology images (\cite{feng2025integrated}). The comprehensive segmentation of multiple targets further facilitates distinguishing diseases that require multi-organ features (e.g., spinal tumors VS endplate infections (\cite{ross2020diagnostic})). 
In addition, TEAMS can be naturally extended to 3D volumetric data and video sequences by using the evolved contours to initialize adjacent slices/frames (\cite{zhao2017}), which exploits the spatiotemporal continuity inherent in these data. This indicates its potential applications in valuable clinical scenarios, such as segmenting critical tissues and surgical instruments from operative videos (\cite{cao2023intelligent}).

\section{Conclusion}
We propose TEAMS (Text-prompted spatiotEmporal dual-heAd Mamba Snake) as a new vision-language Mamba snake framework. TEAMS is characterized by three core designs: (1) Spatiotemporal Snake Evolution Strategy (SSES) modeling bidirectional spatial dependencies among contour points and historical dynamics across evolution steps to accurately predict evolution vectors; (2) Contour Morphology-Aware Mamba (CMAM) modulating point interactions with contour morphology-aware structured state space duality for accurate delineation of fine-grained organ details; and (3) Text-prompted Collaborative {Dual-Head} Snake (TCDHS) incorporating textual cues and leveraging contour information from the evolved snake to mitigate wrong detections. Comprehensive experiments on five challenging medical image datasets of different organs and modalities demonstrate the strong performance of TEAMS in complex multi-organ cases. In summary, this work provides design principles for advancing deep snakes and offers a perspective on Mamba sequence modeling, supporting downstream clinical applications requiring reliable multi-organ boundary {delineation}. 

\section*{Acknowledgements}
This work is supported by the National Natural Science Foundation of China under Grant 62576370, the Foundation for Shenzhen Science and Technology Program under Grants JCYJ20240813151224032 and JCYJ20240813151102004, and the Shenzhen Medical Research Fund under Grant B2402030.

\bibliographystyle{abbrvnat} %
\bibliography{ref}  %

\end{document}